\documentclass[man]{apa7}

\usepackage{lipsum}

\usepackage[american]{babel}

\usepackage{multirow}
\usepackage{graphicx}
\usepackage{subcaption}
\usepackage{csquotes}
\usepackage{comment}
\usepackage[style=apa,sortcites=true,sorting=nyt,backend=biber]{biblatex}
\DeclareLanguageMapping{american}{american-apa}
\title{Deception Detection in Dyadic Exchanges Using Multimodal Machine Learning: A Study on a Swedish Cohort}
\shorttitle{Deception Detection Using Multimodal Machine Learning}

\author{Thomas Jack Samuels$^1$, Franco Rugolon$^2$, Stephan Hau$^1$, Lennart Högman$^1$}
\affiliation{$^1$Department of Psychology, $^2$ Department of Computer and Systems Sciences, Stockholm University, Stockholm, Sweden}

\leftheader{Samuels et al.}

\abstract{This study investigates the efficacy of using multimodal machine learning techniques to detect deception in dyadic interactions, focusing on the integration of data from both the deceiver and the deceived. We compare early and late fusion approaches, utilizing audio and video data—specifically, Action Units and gaze information—across all possible combinations of modalities and participants. Our dataset, newly collected from Swedish native speakers engaged in truth or lie scenarios on emotionally relevant topics, serves as the basis for our analysis.

The results demonstrate that incorporating both speech and facial information yields superior performance compared to single-modality approaches. Moreover, including data from both participants significantly enhances deception detection accuracy, with the best performance (71\%) achieved using a late fusion strategy applied to both modalities and participants. These findings align with psychological theories suggesting differential control of facial and vocal expressions during initial interactions. As the first study of its kind on a Scandinavian cohort, this research lays the groundwork for future investigations into dyadic interactions, particularly within psychotherapy settings.}


\authornote{
   \addORCIDlink{Thomas Jack Samuels}{0009-0008-7181-8666}
   \addORCIDlink{Franco Rugolon}{0000-0002-7693-0576}

  Correspondence concerning this article should be addressed to Thomas Jack Samuels, Department of Psychology, Stockholm University, Albanovägen 12, Stockholm, Sweden
  106 91.  E-mail: thomas.samuels@su.se}

\begin{document}
\maketitle
\subsection{An Introduction to Deception}
Deception has perhaps been most notably defined by \textcite{vrij_detecting_2000} as “a successful or unsuccessful deliberate attempt, without forewarning, to create in another a belief which the communicator considers to be untrue.” In this sense, deception can be conceptualized as a deliberate act involving either the direct manipulation of relevant information or the omission of pertinent detail \parencite{fuller_examination_2012}. The prevalence of these various forms of deception in everyday social interactions, which take place across a range of different human societies \parencite{camara_can_2024}, seems to have been mirrored by increasing academic interest in the topic over the years \parencite{talwar_liar_2022}. The finding that individuals engage in deception behavior more than once a day on average \parencite{depaulo_lying_1996}, highlights the apparent ubiquity and significance of deception in modern society, further reinforcing the need for greater understanding of this important social phenomenon.

Despite the apparent frequency of deceptive behaviors, detecting deception remains a significant challenge for most individuals, with accuracy rates reported at around 54\%, only slightly above chance level \parencite{bond_accuracy_2006}. This difficulty is underscored by the well-documented 'veracity effect', which suggests that lies are detected less accurately than truths \parencite{levine_accuracy_1999}. Furthermore, related cognitive biases, such as 'truth bias', compound the difficulty of accurate deception detection, with individuals generally tending to overestimate the truthfulness of others in social interactions \parencite{zloteanu_tutorial_2024}. Such phenomena emphasize the complexity of deception detection from a human perspective, with inter-individual differences in the propensity to deceive potentially further complicating the task \parencite{eskritt_did_2022}. 

\subsection{Theories of Deception}

Although detecting deception poses challenges for many individuals, the belief that nonverbal cues can serve as useful indicators of truthfulness or deception appears to be particularly pervasive across various cultures and human societies \parencite{bogaard_no_2022}. This belief is rooted in the widespread assumption that nonverbal behaviors are harder to regulate, as they are believed to emerge spontaneously from underlying emotional states \parencite{bogaard_no_2022}. This concept of 'leakage', introduced by \textcite{Ekman_Friesen_1969}, suggests that nonverbal cues emerge as a result of an inability to suppress the oftentimes complex and intense emotional experiences associated with deception. In this framework, nonverbal cues, such as specific facial expressions, bodily poses and other related movements, can be considered as unintentional manifestations of the guilt, anxiety, shame, or even relief, which can occur during deceptive acts. Despite its long-standing influence, the leakage hypothesis \parencite{Ekman_Friesen_1969} has come under significant scrutiny in recent years, with several notable researchers moving away from this theoretical framework \parencite{vrij_reading_2019} in favor of alternative explanations, such as cognitive theories of deception. 

In contrast to the leakage hypothesis, cognitive theories of deception primarily focus on the cognitive demands of deception, with less emphasis on the role underlying emotions in deception situations \parencite{vrij_reading_2019}. In this sense, nonverbal cues are thought to arise, not as leaked emotional responses, but rather as indicators of cognitive load. Deception is conceptualized as being cognitively taxing as individuals attempt to construct coherent fictional narratives, while simultaneously repressing their knowledge of true events \parencite{van_der_zee_liar_2021}. This interplay between maintaining an awareness of the known truth, whilst attempting to replace it with a fabricated untruth, is thought to contribute to cognitive load, which may manifest in subtle nonverbal cues, such as pupil dilation \parencite{van_der_zee_liar_2021}. Moreover, cognitive capacity is further tested, as deceivers attempt to retain a mental representation of both their thoughts and those of the individual(s) they intend to deceive \parencite{zhou_honest_2023}. Within this framework, cognitive theories acknowledge the role of receivers as active participants in deception situations, with related perspectives, such as Interpersonal Deception Theory (IDT), further underscoring the importance of the interactive component in deception \parencite{buller_interpersonal_1996}. 

Interpersonal Deception Theory (IDT) \parencite{buller_interpersonal_1996} outlines that deception is a fundamentally interactive process, involving the dynamic exchange of both verbal and nonverbal information, between senders and receivers \parencite{dunbar_strategic_2020}. Importantly, this theory denotes that senders of deceptive messages actively monitor the reactions of their recipients and adjust their communication based on their perceived degree of believability \parencite{fuller_examination_2012}, thereby framing deception as an inherently interpersonal phenomenon \parencite{eskritt_did_2022}. 

Related theories of deception, such as self-presentational theory \parencite{depaulo_nonverbal_1992}, also focus on the indelibly social nature of deception behavior, with deceivers continually adjusting and refining their social presentation to appear truthful. Like strategic theories of deception, a focus is on how signals (such as nonverbal cues) can be used to strategically convey messages which bolster the facade of truthfulness \parencite{vrij_reading_2019}. 

\subsection{Nonverbal Indicators of Deception}

One of the fundamental principles underpinning and uniting these different theories of deception is the critical role nonverbal indicators play in enabling deceivers to influence their interaction partners' responses. Unsurprisingly, research into nonverbal indicators of deception has steadily increased over the years \parencite{colasanti_did_2024}, with a significant focus on facial, head and bodily movements, as well as prosodic features, including various nonverbal attributes of vocal communication \parencite{Burgoon_Wang_Chen_Pentland_Dunbar_2021}. 

For instance, previous research has provided evidence to suggest that specific facial movements, as categorized using the Facial Action Coding System (FACS) \parencite{ekman_facial_1978}, such as chin raiser (AU17) and lip stretcher (AU20), were associated with deception \parencite{shen_catching_2021, zhou_honest_2023}, whilst lip corner puller (AU12) has been previously found in receivers responding to deceptive senders \parencite{sen_automated_2018}. Other cues, relating to the eyes, such as pupil dilation \parencite{shen_catching_2021} and gaze fixation \parencite{colasanti_did_2024}, have also been associated with deception behavior, whilst specific bodily movements, such as shrugging and fidgeting, have similarly been linked with deception in prior research \parencite{matsumoto_clusters_2021}. 

Prosodic features have also been extensively studied, with research indicating that voices with 'rising intonation, less intensity at the beginning of each syllable, and slower speech rate' were associated with doubt or lying \parencite{goupil_listeners_2021}. These findings align with previous research suggesting that auditory information may be even more useful than visual information in deception detection \parencite{ahmed2024deception}. 

In addition to individual nonverbal cues, considerable research has also examined how the coordination of various nonverbal cues over time, referred to as nonverbal or interactional synchrony, also appears during deceptive interactions. This synchrony has been conceptualized as a social strategy, used by deceivers to enhance their credibility. Indeed, \textcite{duran_conversing_2017}, found an increased rate of head movement coordination in deceptive conversations relative to honest conversations. It has been hypothesized that the reciprocation of nonverbal coordination by a receiver results in an inability to correctly identify and appraise patterns of nonverbal behavior potentially useful for deception detection \parencite{van_der_zee_liar_2021}. As such, deceivers may consciously or non-consciously control the exhibition of nonverbal cues to foster a semblance of credibility, something supported by previous research undertaken by \textcite{burgoon_interpersonal_1994}, which demonstrated that deceivers appeared to display more restrained and tense movements relative to their truthful counterparts. 

Despite the apparent centrality of nonverbal cues in deception, findings in the field appear to demonstrate greater homogeneity than would have previously been expected. This point is perhaps most famously illustrated by a meta-analysis originally conducted by \textcite{depaulo_cues_2003}, which found little evidence to support the role of nonverbal cues in deception. These findings suggest that, if such a relationship does exist, it may be difficult to reliably identify and observe \parencite{delmas_review_2024, vrij_reading_2019}. In this sense, the field lacks a 'Pinocchio's nose' \parencite{vrij_why_2004, Luke_2019}, with some previous research undermined by the use of imprecise measures and potentially inadequate coding systems, leading to mixed findings that offer little support for any single nonverbal feature \parencite{delmas_review_2024}. Such inconsistencies are unsurprising given the influence of broader contextual factors, such as cultural differences, which appear to have a marked impact on the exhibition of nonverbal behaviors \parencite{matsumoto_clusters_2021}. Additionally, some studies have also failed to account for potential confounding variables, such as the distinction made between low vs high-stakes lies, which may further cloud the picture \parencite{Matsumoto_Wilson_2023}. Given these challenges, focusing solely on investigating individual nonverbal cues, or even modalities, may be insufficient \parencite{gupta_bag--lies_2019}, Instead, examining 'clusters' or 'constellations' of cues across multiple modalities may offer a more holistic and nuanced understanding of deception \parencite{hartwig_lie_2014, matsumoto_clusters_2021}. 

\subsection{What is a modality?}
According to \Textcite{baltruvsaitis2018multimodal}, a modality is each of the ways in which a phenomenon happens or is experienced. In particular, a specific group of modalities can be associated with the sensory modalities through which humans experience the world, such as sight, hearing, smell, taste, and tact \parencite{baltruvsaitis2018multimodal}. In this paper, we will focus on visual and vocal signals: visual signals will represent the facial expressions and gaze direction information of the participants, while the vocal signals will represent various parameters associated with the voice of the participants. In addition to modalities, it is important to consider the source of the data, since that will determine, at least in part, the format that the data will have to be analyzed: one of the most relevant distinctions can be made between structured and unstructured data \parencite{zhang2018multi}.
Structured data often originates from forms or sensors, and it refers to information that is organized in a predefined format, or structure: usually it is organized in tables with rows and columns, such as in spreadsheets or databases. Each column has a specific meaning, and each row represents an instance or observation, such as a different subject or point in time. Because of this organization, structured data is straightforward for computers to store, search, and analyze.
In contrast, unstructured data does not follow a clear, predefined format, but can be varies in its contents: it can represent images, text, and other kinds of data. This kind of data is often rich and informative but also more complex to process.

\subsection{Multimodal Machine Learning}
To this end, multimodal  Machine Learning (ML) can be employed. Multimodal ML is a subfield of ML in which data from multiple modalities is integrated and processed, and different algorithms are applied to it \parencite{baltruvsaitis2018multimodal}. The goal in Multimodal ML is to develop models that can make predictions based on data from the different modalities, achieving a more holistic representation of real-world phenomena.

The data from each modality can be integrated at different points of the ML process, but the three main strategies are early fusion, late fusion, and joint/intermediate fusion \parencite{baltruvsaitis2018multimodal,xu2021mufasa}.
\subsubsection{Early fusion}
In early fusion, the data from the different modalities is integrated before applying any ML technique to it, and for this reason, this fusion technique is also called data-level fusion.
Since data from different modalities is usually represented in different ways, each modality is preprocessed independently, with appropriate techniques (e.g., normalization, standardization, tokenization).
After this passage, the data from the different modalities is brought to a common representation. The techniques applied in this step are modality-dependent and vary greatly according to the different modalities that are considered in each experiment.
Once all the data is in a common representation, it can be concatenated, and a single ML model can be applied to the concatenated data to generate predictions \parencite{gadzicki2020early}.

\subsubsection{Late fusion}
Late fusion is also called decision-level fusion: in this process data from the different modalities is preprocessed independently, with appropriate techniques, and a different model is applied to each modality: for example, one model might analyze audio signals, while another might analyze visual data. These models can be as complex or as simple as needed. Once each model has produced the predictions for the data in each modality, these predictions are used to train a meta-model: an ML model that learns to combine the outputs (the predictions) of several other models\parencite{gadzicki2020early}. Meta-models can be seen as a judge that, after listening to the opinions of several experts (the initial models), takes a decision keeping all of these opinions in consideration. In this analogy, the experts do not communicate among them: the initial models are trained only on the data from their specific modality. At the same time, the judge does not see (and would not be able to interpret) the initial data, but only hears the opinions of the experts, and bases their decision only on what the experts tell them.
Meta-models, contrarily to the models that are trained on each modality, have to be simple models to reduce the risk of overfitting: they can be as simple as a majority vote or an average of the decisions of the single models.
The late fusion predictions are the predictions generated by the meta-model.

\subsubsection{Joint fusion}
In joint fusion, the data from each modality is preprocessed independently, and it is fed into a Neural Network (NN), where the fusion step will happen in between the data point and the output point \parencite{gadzicki2020early}. NNs have the ability to process data from different modalities, depending on the kind of layer that is applied to this data \parencite{kahou2013combining}. Moreover, NNs have the ability to highlight non-linear relations in the data and to use these relations to create their predictions, if the correct activation function is used \parencite{grossberg1988nonlinear}. In joint fusion, data from different modalities is fed to different arms of an NN, which applying the appropriate layer to further process the data and bring it to a common representation. Once the data is in a common representation, the representations of the data from each modality can be fused using different techniques, and a common prediction can be outputted by the model \parencite{baltruvsaitis2018multimodal}.

Each fusion strategy has different advantages and disadvantages: early fusion models have the possibility to learn relationships between data from different modalities, but due to the need to have all the data in the same modality \parencite{gadzicki2020early}, some of the characteristics of the data will inevitably be lost, while late fusion, having different models for each modality, is able to exploit all of their characteristics, but since the meta-model is only exposed to the decisions of each base model, the cross-correlations between the different modalities are lost \parencite{gadzicki2020early}.
Joint fusion preserves the ability of the model to learn from each modality independently, and at the same time it preserves, to an extent, the interactions between the different modalities \parencite{baltruvsaitis2018multimodal}, but the models that originate from this fusion strategy are more complex, needing expert crafting, and the risk of overfitting is generally higher. For this reason, these models are not appropriate for tasks with low amounts of data.

\subsection{A Multimodal Understanding of Dyadic Interaction and Deception Detection}

\subsection{ML in deception detection}
Deception detection is a traditionally hard problem, which has been faced using both psychological methods, like inquiring about additional details or asking questions to further investigate a statement, or more advanced ones, like polygraph tests \parencite{prome2024deception}. More recently, ML has been proposed as a tool to help in identifying deception, using behavioral, physiological, and linguistic data \parencite{tang2018resting}.

ML approaches can be classified depending on the modality (or modalities) that they consider in order to predict whether a subject is being deceptive or not: while in past years the use of a single modality was widespread, in recent years approaches that use multiple modalities are gaining traction \parencite{constancio2023deception}: studies using single modalities, such as facial information \parencite{cardaioli2022face, khan2021deception, ahmed2024deception}, textual information \parencite{loconte2023verbal} or body pose information \parencite{poppe2024mining} are becoming rarer, with studies using a combination of modalities, such as facial, audio and textual information \parencite{gogate2017deep}, or facial information and pulse rates \parencite{tsuchiya2023detecting}.
Since ML models ability to detect deception is usually linked to the specific scenario in which the models have been trained, it's important to take these scenarios into consideration: deception can be differentiated according to the importance that it has for the person performing it, in low stakes, medium stakes, and high stakes, where the emotional involvement of the deceiver grows with the stakes \parencite{ahmed2024deception}.
This distinction is important because the ML models do not, usually, detect deception per se, but rather the behavioural changes that originate from a heightened emotional state to infer deception \parencite{constancio2023deception}.
In this scenario, synthetic datasets, which are commonly used to train ML models in deception detection, and which also usually have the highest number of available samples, usually lack the high stake component that real life datasets can carry.

Multiple studies use databases created in unnatural conditions, in which the subjects do not interact with other humans, to train their models, such as the SMIC dataset \parencite{li2013spontaneous}, CASME \parencite{yan2013casme}, CASME II \parencite{yan2014casme}, SASE-FE \parencite{wan2017results}, while other studies used datasets from real life situations, such as the real-life trial dataset \parencite{perez2015deception}, but rich, multimodal, annotated real-life datasets are rare.
Multiple studies \parencite{prome2024deception, constancio2023deception, khan2021deception, ahmed2024deception} cite the importance of micro-expressions \parencite{ekman2009telling}, fleeting variations of one's facial expression as possible indicators of deception.
To capture such expressions, it's important to consider the temporal component of the data, analyzing the changes in the facial expressions of the subjects, but there are also papers \parencite{cardaioli2022face} which fail to do so, employing techniques such as averaging the activation of the different AUs over a period of time, losing the temporal dimension of the data and the ability to capture rapid changes.

\subsection{Problem definition}
The general task in this experiment was to determine whether the sender was lying or telling the truth during their interaction with a passive receiver. We collected audio and video recordings of participants in the deception experiment that we designed, and we extracted Facial Action Units (FAC) activation levels and eye movements from each frame of each video recording, and the Geneva minimalistic acoustic parameter set (GeMAPS) \parencite{eyben2015geneva} from each timepoint in the acoustic recordings.
We then aggregated the recordings of all the dyads for each modality into a collection of files containing the features extracted from the video recordings of the sender in all the dyads, and into a separate files collection containing the same features for both participants.
We then repeated the process with the features extracted from the audio recordings, to obtain files containing the data extracted from the recordings of the sender, and a different set of files containing the data extracted from the recordings of both participants.
We then analyzed these files using three different approaches: the unimodal approach, the early fusion approach, and the late fusion approach.
Given the low amount of data available in our dataset, it was not possible to utilize the joint fusion approach, since the risk of overfitting, with sucha  complex model, would have been too high.

In the unimodal approach, we considered either the video or audio recordings of the active participant or both participants in all the dyads.

Our objective was to train a classifier using leave-one-out cross-validation \parencite{han2022data} at the dyad level, which, when applied separately to either facial or vocal features extracted from recordings of the sender alone or from both participants, would yield a binary prediction (``lie'' or ``truth'') for each dyad.
We then averaged the results of the leave-one-out cross-validation to get the final precision and recall metrics per each class, and the final accuracy metric.

In the early fusion approach, we considered both video and audio recordings of the same dyads involved in the unimodal approach. Since the audio and video recordings had different recording frequencies, our goal was to bring them to the same number of timesteps before learning a classifier with the same procedure used in the unimodal approach. In particular, we experimented with two different methods to bring the extracted features to the same feature space, using either Piecewise Aggregate Approximation (PAA) \parencite{keogh2001dimensionality} or Symbolic Aggregate approXimation (SAX) \parencite{lin2007experiencing} to bring the recordings of both modalities to have the same amount of timesteps. We applied two commonly used timeseries summarization techniques, PAA and SAX. These methods both apply a segmentation technique, calculating the average value of the timeseries for the each of the resulting segments. On top of this segmentation and averaging technique, SAX applies a discretization procedure to reduce the complexity of the values for each timestep.

In the Late fusion approach we used a two-step process, learning the base classifiers with the same procedure as for the unimodal approach, but, in addition, we trained a decision tree on the predictions of the base classifiers for the dyads that were not the target of the classification during the leave-one-out cross-validation procedure, and we used this decision tree to produce the final predictions of ``lie'' or ``truth'' for each dyad.

\section{Method}
In our approach, we analyzed both visual (FAC and gaze information extracted using OpenFace) and audio (GEMAPS features extracted using OpenSmile) information from participants who were assigned the task of lying or telling the truth. 

\subsection{Participants}
44 individuals participated in this experimental study, with an age range from 18 to 67 years old (M = 29.52, SD = 11.2). The sample contained 28 men, 14 women and 2 identifying as 'other'. All participants were native Swedish speakers and were at least 18 years old at the time of participation. Additional details relating to the sample for each experimental condition can be found in \ref{tab:table_one}.

\begin{table}[t]
\centering
\caption{Summary statistics for the age, gender and education level of the participants in the study.}
    \label{tab:table_one}
\resizebox{0.75\textwidth}{!}{%
\begin{tabular}{llllll}
\toprule
                          &                             & Overall     & Lie        & Truth       \\
\hline
 n                        &                             & 44          & 20         & 24          \\
 Age, mean (SD)           &                             & 29.1 (11.1) & 30.4 (8.8) & 28.0 (12.9) \\
 Age Groups, n (\%)       & 18-30                       & 30 (68.2)   & 11 (55.0)  & 19 (79.2)   \\
                          & 31-40                       & 8 (18.2)    & 7 (35.0)   & 1 (4.2)     \\
                          & 41-50                       & 4 (9.1)     & 2 (10.0)   & 2 (8.3)     \\
                          & 51-60                       & 1 (2.3)     &            & 1 (4.2)     \\
                          & 61-70                       & 1 (2.3)     &            & 1 (4.2)     \\
Gender, n (\%)            & Female                      & 28 (63.6)   & 13 (65.0)  & 15 (62.5)   \\
                          & Male                        & 14 (31.8)   & 7 (35.0)   & 7 (29.2)    \\
                          & Other                       & 2 (4.5)     &            & 2 (8.3)     \\
 Education Level, n (\%)  & Middle school               & 3 (6.8)     & 1 (5.0)    & 2 (8.3)     \\
                          & High School                 & 16 (36.4)   & 7 (35.0)   & 9 (37.5)    \\
                          & Vocational training         & 3 (6.8)     & 2 (10.0)   & 1 (4.2)     \\
                          & Bachelor Degree             & 14 (31.8)   & 7 (35.0)   & 7 (29.2)    \\
                          & Post graduate qualification & 8 (18.2)    & 3 (15.0)   & 5 (20.8)    \\
                          
\bottomrule
\end{tabular}}
\end{table}

Participants were recruited through the use of several mediums, including advertisements posted on digital learning platforms and bulletin boards located throughout Stockholm University, as well as on the research recruitment platform Accindi (accindi.se). All participants received financial compensation in the form of a 100 SEK gift card for a Swedish supermarket chain. 

\begin{figure}
    \centering
    \includegraphics[width=0.75\linewidth]{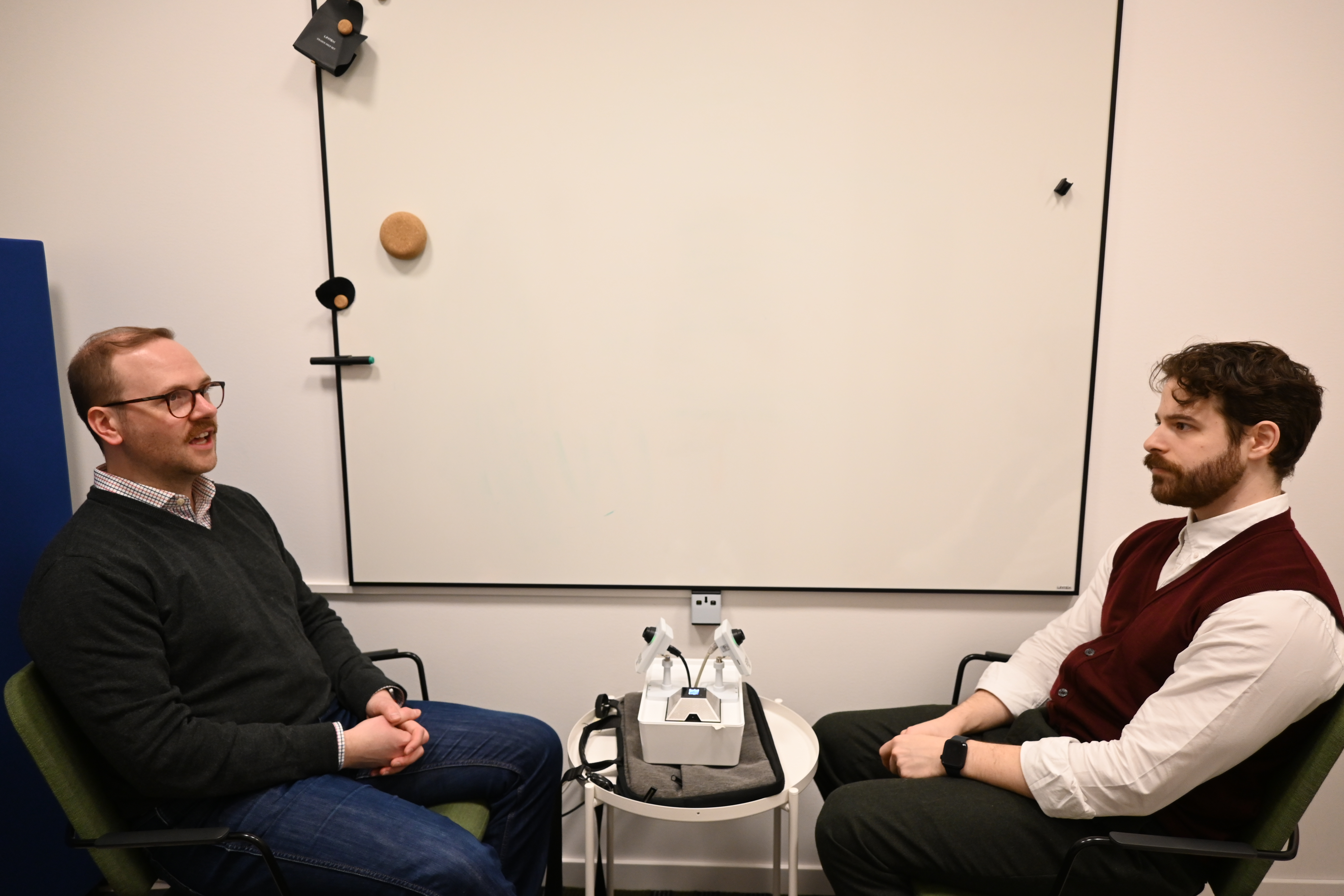}
    \caption{A photograph showing the experimental setup, with the two cameras in the center and two participants taking part in the experimental interaction.}
    \label{fig:setup}
\end{figure}
\subsection{Procedure}
The study employed a between-subjects design, with participants randomly assigned to experimental dyads, in one of two experimental conditions: truthfulness or deception. Within each dyad participants were assigned the role of either a) sender or b) responder. In the truth-telling condition, person A, the sender, was advised to recall actual experiences, completing seven sentences, such as 'One of my most cherished memories is when...', and 'A major problem I’ve had was when...'. In the deception condition, the sender was asked to complete the same sentences but with fabricated responses which did not reflect their true experiences, beliefs or opinions. Participants adopting the role of the sender in both conditions were advised to rehearse their statements before attending the experimental session, in an attempt to minimize any differences potentially occurring due to familiarity with the experience or opinion \parencite{depaulo_cues_2003}. Participants assigned to the role of responder were tasked to ask follow-up questions after each sentence was read, in an attempt to enter into a meaningful conversation for each statement. 

Before the experiment began, the experiment leaders ensured that all participants had been appropriately prepared and provided consent to participate. Participants were then instructed to remove any items which obscured their face or head (e.g. caps or scarves) and remove any oral impediments (e.g. nicotine pouches or food items, such as chewing gum), before being seated in front of their respective cameras. Each participant was recorded with a separate camera which faced directly toward them, as shown in Picture \ref{fig:setup}. 

Participants were then introduced to one another and instructed to engage in conversation for approximately one minute, during which the experimental leaders ensured that the recording equipment was functioning correctly and that the cameras and microphone were capturing the appropriate audiovisual input. Once these preliminary checks were complete, the experiment leader instructed the participants to begin, with the sender reading their first pre-prepared statement. The responder then asked a follow-up question, with conversation unfolding naturally after this point. After two minutes, the experiment leader instructed the sender to move on to their second statement, and the process was repeated. In total, there were seven rounds of conversation, one per pre-prepared statement, each lasting two minutes, with the total interaction time of being around 15 minutes. 

At the end of the interaction, participants were partitioned by a screen and asked to complete a set of questionnaires, rating the quality of their interaction and the perceived truthfulness of their interaction partner. After completing the questionnaires, participants were reunited and debriefed. They were then reimbursed with gift cards and advised to contact the lead researcher with any follow-up questions. 

\subsection{Materials}
Participants were required to complete an informed consent form before participating in the experiment. The form provided a brief description of the structure of the experiment, as well as relevant information relating to data management and storage. 

Post-experimental questionnaires were used to gather additional data. These included a perceived trustworthiness scale, a three-item scale adapted and translated (from English to Swedish) from the scale originally used by \textcite{Dunbar_Giles_Bernhold_Adams_2019} ($\alpha$ = .81). This scale assessed participants' perceptions of the truthfulness of their interaction partners, with responses recorded on a 7-point Likert scale. Participants also completed a subset of three items taken from the original scale presented by \textcite{bernieri_dyad_1996} ($\alpha$ = .83), evaluating the overall experience of the interaction. The items asked participants to rate how 'engaging', 'cooperative', and 'awkward' they perceived the interaction to be, with the scale translated into Swedish.

\begin{figure}
\centering
\begin{subfigure}{0.37\textwidth}
\centering
\includegraphics[width = \textwidth]{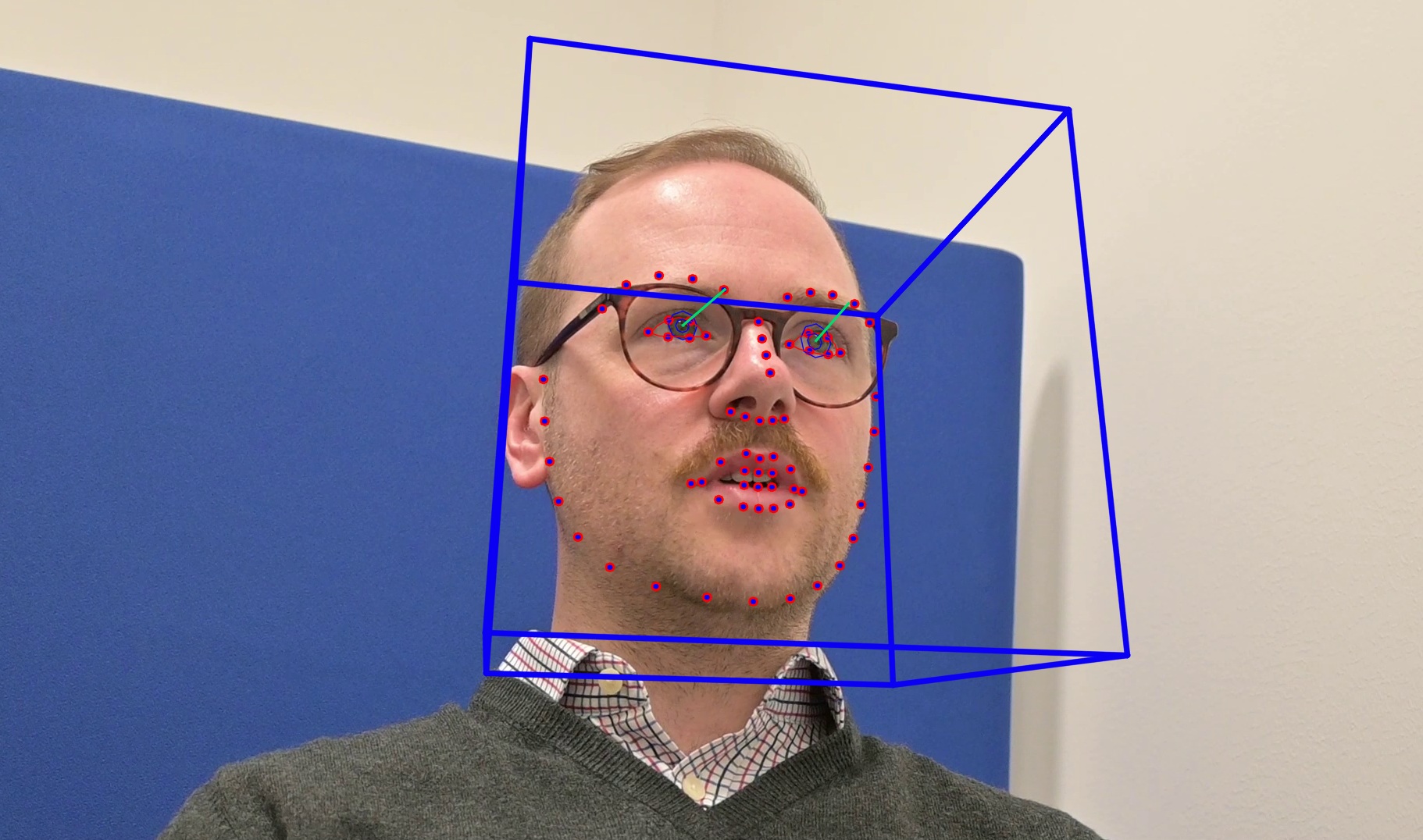}
\end{subfigure}
\begin{subfigure}{0.37\textwidth}
\centering
\includegraphics[width = \textwidth]{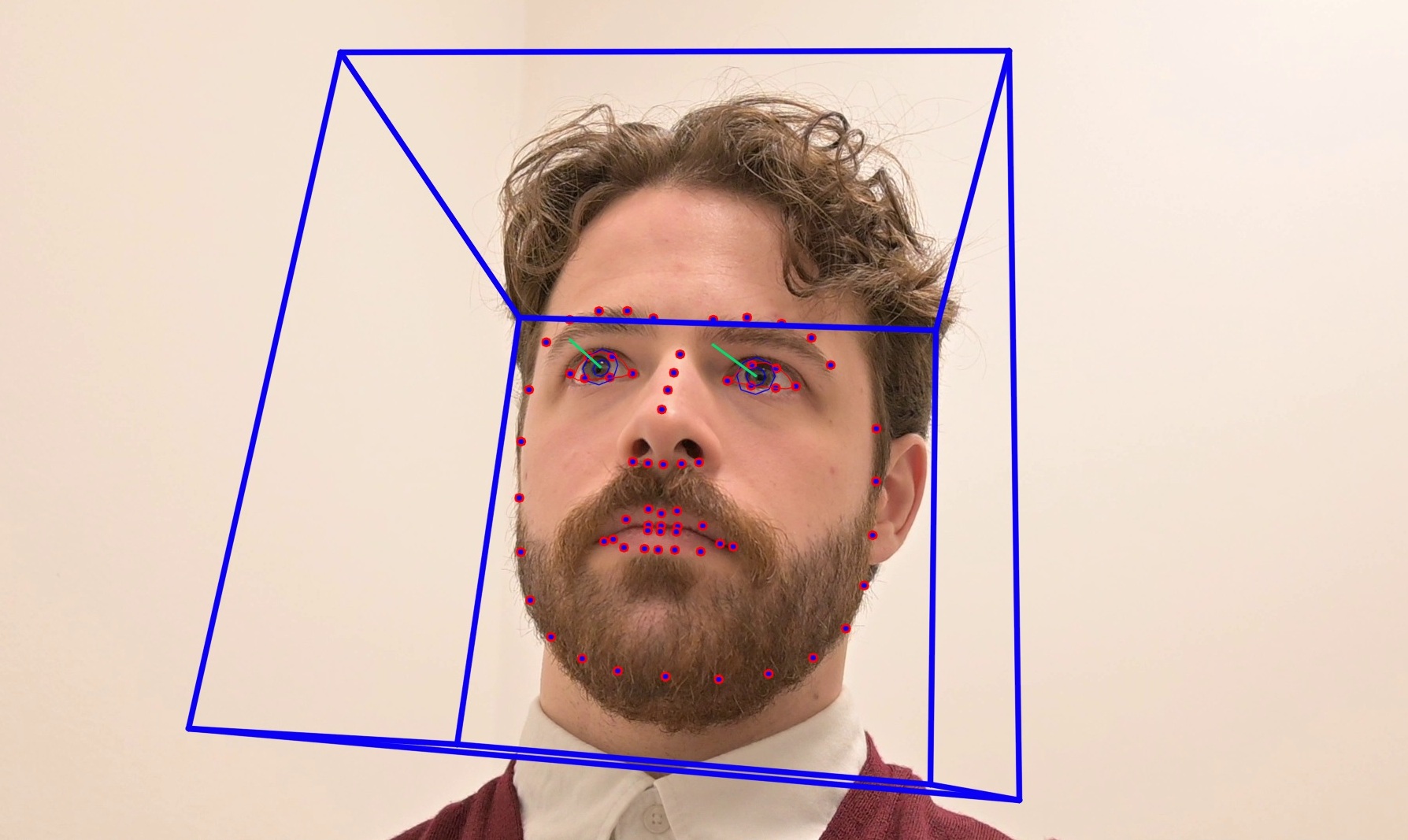}
\end{subfigure}
\caption{A screenshot showcasing the extraction process for facial features from the video recordings.}
 \label{fig:openface_combined}
\end{figure}

\subsection{ML methods}
After the experiments were conducted, we collected the video and audio recordings of the sessions, and we used two open source programs, OpenFace \parencite{baltruvsaitis2016openface} and OpenSmile \parencite{eyben2010opensmile}, to automatically extract features from them.
OpenFace allows us to extract AU and information about the eye position and gaze angles, collectively called facial features, while we use OpenSmile with the GeMAPS \parencite{eyben2015geneva} configuration to extract acoustic features.

The facial features are extracted for each frame of the recording, and are concatenated on a temporal axis to create time series for each feature.
In the same way, the acoustic features are extracted from the recordings, extracting a hundred time points per second, and concatenated along the temporal axis to create time series for each feature.

After the concatenation, the time series relative to each modality for each participant were transformed into two-dimensional arrays (features $\times$ timesteps).
The timeseries of the different participants were then aggregated for each modality, to create a three-dimensional array (participants $\times$ features $\times$ timesteps) that was modality specific.
The features contained in these three-dimensional arrays were then standardized feature-wise and participant-wise using z-score normalization \parencite{han2022data}, to reduce the impact of the use of different measuring units on the classification process.

We divided the classification process, as described in the problem formulation subsection, in three phases: unimodal, early fusion, and late fusion.
In the unimodal phase, we trained three classifiers, namely Rocket \parencite{dempster2020rocket}, Canonical Interval Forest \parencite{middlehurst2020canonical} and Z-Time \parencite{lee2024z} on the data from each modality. We repeated the process using data from the sender only and from both participants. To reduce the risk of overfitting, we employed leave-one-out cross-validation, using the data from $n-1$ participants or dyads as our training set and the data from the remaining participant or dyad as our test set, and repeating the process $n$ times, until the data from each participant or dyad was used as test data exactly once.
We repeated this process for both modalities, therefore using both data from the facial features and the acoustic features, separately, to predict the binary label to be attributed to the data.

In the early fusion phase, as described in the introduction, the data from different modalities has to be brought to a common space, concatenated, and then a classifier can be applied to the concatenated data.
Since both modalities were made of timeseries, albeit with different dimensionalities ($22 dyads \times 19739 timesteps \times 25 features$ for the facial features of each participant, and $22 dyads \times 66105 timesteps \times 24 features$ for the acoustic features of each participant) we needed to bring the different timeseries to the same number of $timesteps$. To do so, we compared two different techniques: PAA \parencite{keogh2001dimensionality} and SAX \parencite{lin2007experiencing}. Given an original timeseries $t$ of $N$ timesteps, both techniques allow us to choose a desired number $N'$ of frames, where $N' \leq N$. They then divide the original timeseries in $N'$ fragments of the same duration, to then calculate the average value of each fragment. This generates a new timeseries $t'$ with the desired duration $N'$. In addition to this, SAX also applies a discretization over the values $V$ of the timeseries $t$, dividing the interval between the minimum value $V_{min}$ and the maximum value $V_{max}$ in $V'$ desired values, and creating a dictionary of $V'$ values. The original values $V$ are then mapped to the new $V'$ values, reducing the complexity of the timeseries.

In our early fusion approach, we chose 10000 timesteps for both PAA and SAX, but, without loss of generality, any $N'$ which is less or equal to the duration of both timeseries can be chosen. After applying PAA and SAX, the facial features were represented by a three-dimensional array of $22 dyads \times 10000 timesteps \times 25 features$, or $50 features$ if considering the data of both participants, and the acoustic features by a three-dimensional array of $22 dyads \times 10000 timesteps \times 24 features$, or $48 features$ if considering the data of both participants, allowing us to concatenate the features at each time point. After concatenation, a new three-dimensional array of dimensions $22 dyads \times 10000 timesteps \times F features$ was generated. The actual number of features depended on the use of data from just the sender or both participants for each modality.

Following the same training and testing procedure used in the unimodal phase, we trained a Rocket binary classifier across the entire participant dataset, systematically evaluating all possible combinations of modalities, and using data from either only the sender or both participants for each modality. Furthermore, we used either PAA and SAX to bring the timeseries from different modalities to the same feature space. Again, we used leave-one-out cross-validation to reduce the risk of overfitting.

In the late fusion phase, as explained in the introduction, the data from each modality is fed to a classifier specific for that modality, and then the predictions of each classifier are either aggregated or fed to a meta-classifier.
In our case, we used the probabilistic predictions generated by Rocket on the single modalities, and we used those to train a decision tree and produce the final multimodal predictions.

To do this, we, once more, used leave-one-out cross-validation to reduce the risk of overfitting. To further reduce the risk of overfitting, we kept three as the maximum depth of the tree.

The primary aim of this study was to explore the impact of different modalities, combinations of modalities, and of the dyadic aspect (i.e., using data from the sender, or from the sender and the receiver) on deception detection performance. As such, our focus was on comparative analysis across experimental conditions rather than on maximizing the absolute predictive performance of any single classifier. Consequently, we did not perform hyperparameter tuning for our models. All base classifiers were trained using their default parameter settings to ensure consistency and comparability across conditions, and the decision tree meta-classifier was trained using a reduced maximum depth to reduce the risk of overfitting and to isolate the effects of modality and data origin.

Since there was only a minimal imbalance in the class distribution in the data, we deemed that accuracy was a good performance measure for the classifiers. To give a better idea of the performance of each classifier on each class, we still reported precision and recall per each class for all the classifiers \footnote{All the code needed to run these experiments is made available on \href{https://github.com/FoxtrotRomeo/DeceptionDetection}{GitHub}.}.

\section{Results}
Table~\ref{tab:results} summarizes the performance of the different classifiers on the various combinations of modalities and participants for this study.

Looking at the results for the unimodal classifiers, we can see that in both single modalities, employing data from both participants either does not change the accuracy of the classifier in two cases, or improves it in the other four cases.
For facial expressions, in particular, there is an increase in accuracy using data from both participants, but even the best results are only marginally better than random level.
We can also see that sensitivity always improves, while the results for specificity are mixed.

Looking at the results for the early fusion, we can see that applying SAX instead of PAA improves the classification accuracy and the sensitivity in three out of four modality combinations, while using data from the facial expressions of both participants improved the classification accuracy in four out of eight cases, decreasing it in the remaining half, compared to using data from the facial expressions of just the sender, and using data from the voice of both participants improved the classification accuracy in two cases out of eight, decreased it in two other cases, and left it unchanged in the remaining half.

Looking at the results for the late fusion, we can see that this way to perform data fusion generated the two best classification accuracies of the whole experiment, in both cases using data from the facial expressions of the sender. The best accuracy overall was obtained by the decision tree trained on the output of the Rocked classifiers trained on the voices of both participants, and the facial expressions of the sender. We can also see that this classifier shows a perfect precision for class 0 (lying dyads) and a perfect recall for class 1 (truthful dyads), meaning that this classifier was able to spot all the dyads in which the sender was telling the truth, and never misclassified a dyad in which the sender was lying.

Finally, looking at the results for human classification, the receivers showcased a lower accuracy than unimodal methods trained on prosodic data, but a better accuracy than unimodal methods trained on facial expressions. The early fusion approach showed better results than the receivers in three experimental combinations out of eight, in all three cases when using data from both participants in at least a modality and when using SAX for dimensionality reduction. Late fusion showed better results than the receivers in half of the experimental settings.

\begin{table}[]
\centering
\caption{A table displaying the performance of the different classifiers on both unimodal and multimodal data. The best result for each metric is highlighted in \textbf{bold}, and the accuracy results that are at least human-level are highlighted with \underline{underlined text}.}
\label{tab:results}
\resizebox{\textwidth}{!}{%
\begin{tabular}{cccccccc}
\toprule
Modality &
  Participants &
  Method &
  \multicolumn{5}{c}{Performance} \\ \cline{4-8} 
 &
   &
   &
  \begin{tabular}[c]{@{}c@{}}Precision\\ Class Lie\end{tabular} &
  \begin{tabular}[c]{@{}c@{}}Precision\\ Class Truth\end{tabular} &
  \begin{tabular}[c]{@{}c@{}}Recall\\ Class Lie\end{tabular} &
  \begin{tabular}[c]{@{}c@{}}Recall\\ Class Truth\end{tabular} &
  Accuracy \\ \hline
\multirow{6}{*}{Voice} &
  \multirow{3}{*}{Sender} &
  Rocket &
  0.58 &
  0.70 &
  \textbf{0.70} &
  0.58 &
  \underline{0.64} \\
 &
   &
  CIF &
  0.62 &
  0.64 &
  0.50 &
  0.75 &
  \underline{0.64} \\
 &
   &
  Z-time &
  0.60 &
  0.67 &
  0.60 &
  0.67 &
  \underline{0.64} \\ \cline{2-8} 
 &
  \multirow{3}{*}{Both} &
  Rocket &
  0.67 &
  0.69 &
  0.60 &
  0.75 &
  \underline{0.68} \\
 &
   &
  CIF &
  0.67 &
  0.62 &
  0.40 &
  0.83 &
  \underline{0.64} \\
 &
   &
  Z-time &
  0.67 &
  0.69 &
  0.60 &
  0.75 &
  \underline{0.68} \\ \hline
\multirow{6}{*}{\begin{tabular}[c]{@{}c@{}}Facial\\ Expressions\end{tabular}} &
  \multirow{3}{*}{Sender} &
  Rocket &
  0.25 &
  0.43 &
  0.20 &
  0.50 &
  0.36 \\
 &
   &
  CIF &
  0.29 &
  0.47 &
  0.40 &
  0.58 &
  0.41 \\
 &
   &
  Z-time &
  0.29 &
  0.47 &
  0.20 &
  0.58 &
  0.41 \\ \cline{2-8} 
 &
  \multirow{3}{*}{Both} &
  Rocket &
  0.50 &
  0.57 &
  0.40 &
  0.67 &
  0.55 \\
 &
   &
  CIF &
  0.43 &
  0.53 &
  0.30 &
  0.67 &
  0.50 \\
 &
   &
  Z-time &
  0.20 &
  0.47 &
  0.10 &
  0.67 &
  0.41 \\ \hline
\multirow{8}{*}{\begin{tabular}[c]{@{}c@{}}Multimodal\\ Early Fusion\end{tabular}} &
  \multirow{2}{*}{\begin{tabular}[c]{@{}c@{}}Voice: Sender\\ Facial Expressions: Sender\end{tabular}} &
  Rocket with PAA &
  0.50 &
  0.56 &
  0.30 &
  0.75 &
  0.55 \\
 &
   &
  Rocket with SAX &
  0.44 &
  0.54 &
  0.40 &
  0.58 &
  0.50 \\ \cline{2-8} 
 &
  \multirow{2}{*}{\begin{tabular}[c]{@{}c@{}}Voice: Sender\\ Facial Expressions : Both\end{tabular}} &
  Rocket with PAA &
  0.50 &
  0.58 &
  0.50 &
  0.58 &
  0.55 \\
 &
   &
  Rocket with SAX &
  0.57 &
  0.60 &
  0.40 &
  0.75 &
  \underline{0.59} \\ \cline{2-8} 
 &
  \multirow{2}{*}{\begin{tabular}[c]{@{}c@{}}Voice: Both\\ Facial Expressions: Sender\end{tabular}} &
  Rocket with PAA &
  0.29 &
  0.47 &
  0.20 &
  0.58 &
  0.41 \\
 &
   &
  Rocket with SAX &
  0.62 &
  0.64 &
  0.50 &
  0.75 &
  \underline{0.64} \\ \cline{2-8} 
 &
  \multirow{2}{*}{\begin{tabular}[c]{@{}c@{}}Voice: Both\\ Facial Expressions: Both\end{tabular}} &
  Rocket with PAA &
  0.50 &
  0.58 &
  0.50 &
  0.58 &
  0.55 \\
 &
   &
  Rocket with SAX &
  0.56 &
  0.62 &
  0.50 &
  0.67 &
  \underline{0.59} \\ \hline
\multirow{8}{*}{\begin{tabular}[c]{@{}c@{}}Multimodal\\ Late Fusion\end{tabular}} &
  \multirow{2}{*}{\begin{tabular}[c]{@{}c@{}}Voice: Sender\\ Facial Expressions: Sender\end{tabular}} &
  \multirow{2}{*}{Decision Tree} &
  \multirow{2}{*}{0.75} &
  \multirow{2}{*}{\textbf{0.71}} &
  \multirow{2}{*}{0.60} &
  \multirow{2}{*}{0.83} &
  \multirow{2}{*}{\underline{0.73}} \\
 &
   &
   &
   &
   &
   &
   &
   \\ \cline{2-8} 
 &
  \multirow{2}{*}{\begin{tabular}[c]{@{}c@{}}Voice: Sender\\ Facial Expressions: Both\end{tabular}} &
  \multirow{2}{*}{Decision Tree} &
  \multirow{2}{*}{0.40} &
  \multirow{2}{*}{0.50} &
  \multirow{2}{*}{0.40} &
  \multirow{2}{*}{0.50} &
  \multirow{2}{*}{0.45} \\
 &
   &
   &
   &
   &
   &
   &
   \\ \cline{2-8} 
 &
  \multirow{2}{*}{\begin{tabular}[c]{@{}c@{}}Voice: Both\\ Facial Expressions: Sender\end{tabular}} &
  \multirow{2}{*}{Decision Tree} &
  \multirow{2}{*}{\textbf{1.00}} &
  \multirow{2}{*}{\textbf{0.71}} &
  \multirow{2}{*}{0.50} &
  \multirow{2}{*}{\textbf{1.00}} &
  \multirow{2}{*}{\underline{\textbf{0.77}}} \\
 &
   &
   &
   &
   &
   &
   &
   \\ \cline{2-8} 
 &
  \multirow{2}{*}{\begin{tabular}[c]{@{}c@{}}Voice: Both\\ Facial Expressions: Both\end{tabular}} &
  \multirow{2}{*}{Decision Tree} &
  \multirow{2}{*}{0.43} &
  \multirow{2}{*}{0.53} &
  \multirow{2}{*}{0.30} &
  \multirow{2}{*}{0.67} &
  \multirow{2}{*}{0.50} \\
 &
   &
   &
   &
   &
   &
   &
\\ \hline
  \multirow{2}{*}{Receiver} &
\multirow{2}{*}{\begin{tabular}[c]{@{}c@{}}Voice: Sender\\ Facial Expressions: Sender\end{tabular}} &
  \multirow{2}{*}{Personal experience} & 
  \multirow{2}{*}{\textbf{1.00}} &
  \multirow{2}{*}{0.57} &
  \multirow{2}{*}{0.10} &
  \multirow{2}{*}{\textbf{1.00}} &
  \multirow{2}{*}{\underline{0.59}} \\

  &
   \\ \bottomrule
\end{tabular}%
}
\end{table}

\section{Discussion}

Overall, the present findings indicate that adopting a multimodal approach to deception detection can be advantageous, as shown by a trend in recent research \parencite{gogate2017deep, tsuchiya2023detecting} especially when data obtained from different nonverbal modalities are integrated using late fusion. One particularity of this study is that all the data comes from non-verbal modalities, decreasing the need for extremely complex models such as Large Language Models, which would be less interpetable and would require more computational power to run. Additionally, the data for this study can be collected in a relatively non-invasive way, with cameras and microphones that can be located away from the subjects, reducing the likelihood for the recording procedure to influence the behavior of the subjects. While the use of pulse rates, as was done by \textcite{tsuchiya2023detecting} or skin conductivity \parencite{strofer_deceptive_2015} can provide information on physiological signals that are more difficult to control for the sender, it can also disrupt the normal flow of the interaction. Since, in this study, we wanted to focus on the dyadic aspect of the interaction, we chose to only use audio and video information. Moreover, even in the video information, we were limited in the type of video information by the positioning of the cameras: as it can be seen in Picture \ref{fig:openface_combined}, our cameras only captured the face of the participants and, partially, the upper part of the shoulders, rendering the extraction of body positions impossible.

Even with these limitations, Within the current study, the highest accuracy (0.77) was significantly higher than what humans usually are able to achieve.

One particularly notable result from our experiments is the perfect precision achieved for the Lie class in our best-performing model. Precision, in this context, measures the proportion of instances classified as lies that are actually lies. A perfect precision score of 1.00 means that every time the model labeled a statement as a lie, it was correct: there were no false accusations. This is especially important in deception detection tasks, where false positives can carry serious ethical and practical consequences. In forensic, clinical, or even interpersonal settings, wrongly labeling someone as deceptive can damage trust, reputation, or bring legal outcomes. Therefore, a system that avoids such errors is particularly valuable.

This result was achieved using a Decision Tree model as a meta-model, trained on the outcomes from two Rocket classifiers, which, in turn, were trained on the data from the single modalities, audio and video data. This approach suggests that such late fusion techniques may enable the single modality classifiers to learn unique, complementary information obtained from each modality, which, when combined at a later decision stage, can lead to greater classification accuracy.
This advantage seems to outweigh the lack of interactions among the single modalities, which might not be determinant for a good classification result, as this approach does not preserve them.
Additionally, the late fusion approach allowed us to maintain the data from each modality in its original format, avoiding the issue of having to conduct dimensionality reduction, which we had to face in the case of early fusion.

In contrast, the use of individual modalities within this study yielded lower accuracy scores. The use of facial expression data alone produced lacklustre returns, with accuracy rates ranging from 0.36 - 0.55, indicating that the use of such information alone may not be sufficient for accurate classification of deception in such contexts. Indeed, data relating to the voice proved to produce marginally better results, although these also fell short of the scores produced by the late fusion multimodal model.

To use early fusion, it was fundamental to reduce the timeseries from different modalities to a common, lower dimensionality. This process was carried out with two different methods, using the same classifier, to compare the impact of the dimensionality reduction techniques. In three out of four cases where the two techniques, PAA and SAX, were compared, the results of the classifier trained on the data generated by SAX were superior than those of the results of the classifier trained on data transformed with PAA. Moreover, the results of the classifiers trained on early fusion data were worse than the results of the classifiers trained using only the voice of both participants, showing that the direct integration of data from both modalities is not necessarily beneficial during the classification process.

The present results may be tentatively interpreted through the frame of existing theories of deception. The relatively poor performance of classifiers trained solely on facial expression datav raises important questions about the robustness of leakage theory's emphasis on the face as the most primary channel through which involuntary emotional cues may 'leak' \parencite{Ekman_Friesen_1969}. If such leakage were a consistent or detectable phenomenon, we may expect to see a markedly better performance in terms of classification accuracy for this individual modality. Such results align with previous findings \parencite{depaulo_cues_2003}, and may even suggest that such features may be more subject to voluntary control than previously assumed, an idea originally outlined by \textcite{burgoon_interpersonal_1994}. 
From the cognitive perspective, an increase in cognitive load may not be uniformly distributed across modalities, but instead disproportionately affect those modalities which place high demands on temporal and semantic coordination, such as speech \parencite{depaulo_cues_2003}. The stronger performance of classifiers trained on prosodic features aligns with this perspective, suggesting this modality may be more susceptible to disruption under cognitive load. Indeed, the improved performance of the late fusion model over the early fusion model may further support the notion that deception cues are not uniformly distributed across modalities, but instead emerge unevenly throughout the dynamic and shifting flow of a nonverbal exchange. This pattern aligns with Interpersonal Deception Theory (IDT) \parencite{buller_interpersonal_1996}, which posits  that deception is a multimodal, dynamic process evolving across the course of an interaction. 

Such findings appear to be consistent with previous research which has outlined the benefits of using a multimodal approach in the detection of deception, as well as the potential pitfalls of unimodal classification \parencite{camara_can_2024}. For example, previous research on the analysis of single modalities, such as facial expressions, has found evidence to suggest that such displays may be highly idiosyncratic, with previously unexamined factors, such as personality, potentially playing an important role in explaining underlying variance apparently present in these nonverbal behaviors \parencite{zhou_honest_2023}. Furthermore, the examination of single modalities, often presented as single nonverbal channels, may not fully capture the wealth of information being communicated across multiple nonverbal channels simultaneously, leading to an inability for models to correctly classify and thereby detect deception \parencite{matsumoto_clusters_2021}.
Furthermore, the performance comparison between human and ML deception detection confirms the promise of ML in this delicate domain, which might be due to the absence of truth biases showcased, in previous research, by human participants \parencite{zloteanu_tutorial_2024}. 
Although deception can be better detected from multiple cues than a single cue, the strongest cue contributes a lot. 

However, multimodal deception detection is not a panacea, and the application of such approaches should be carefully considered, especially when the considering the possiblity of Type I errors in not correcting for multiple comparisons \parencite{Luke_2019}. The current approach employs a leave-one-out cross-validation in an attempt to reduce overfitting on the current sample. In future, authors may seek to further reduce the likelihood of Type I errors by addressing these issues earlier in the analytical workflow.

This study constitutes an illustration of a method that can be implemented within the burgeoning research field of multimodal deception detection and multichannel nonverbal research in general \parencite{Matsumoto_Wilson_2023}. As such current study utilizes a sample of 22 dyads, comprising of 44 individuals who were previously unknown to one another. Whilst previous research has featured samples of a similar size with sufficient statistical power to detect effects \parencite{zhou_honest_2023}, we are mindful to not overstate our findings given previous criticism of sample sizes in this field \parencite{Luke_2019}. Whilst leave-one-out cross-validation can reduce the likelihood of overfitting in smaller samples, it may still not appropriately address more foundational concerns such as the representativeness of the sample, something which future research could seek to address. 

Furthermore, whilst the use of a laboratory setting enabled us to establish a stable ground-truth for this study, through the use of experimental manipulation \parencite{Dunbar_Burgoon_Chen_Wang_Ge_Huang_Nunamaker_2023}, the generalizability of our findings may be limited by the use of an artificial setting which could be viewed as a somewhat unnatural context for the study of human interaction patterns \parencite{Burgoon_Wang_Chen_Pentland_Dunbar_2021}. In addition, although the pairing of strangers helped control for potential familiarity effects in nonverbal interaction patterns \parencite{Brandt_Miller_Hocking_1980b}, it may also have reduced the perceived stakes of engaging in deception, with interactants unlikely to engage in future interactions, thereby lessening the potential consequences of being caught in a lie when compared to real-world scenarios \parencite{Dunbar_Giles_Bernhold_Adams_2019}. Such 'low-stakes' have long been assumed to result in fewer nonverbal deceit cues when compared to 'high-stakes' scenarios which could induce stronger emotional responses and result in greater cognitive load \parencite{vrij_reading_2019}. However, given that participants were asked to fabricate personal experiences or opinions, it remains an open question whether the potentially emotionally-charged nature of these statements could have, to some degree, heightened the stakes of deception within our study. Additionally, our experimental method seeks to instigate a relatively narrow form of deception, namely falsified experiences or beliefs, and does not account for broader definitions of deception which may include related phenomena such as gas-lighting, feigned ignorance and self-deception. However, adopting this simplified definition of deception did enable us to establish a clear ground truth and verify the effectiveness of the experimental manipulation through post-interaction self-reports. 

\subsection{Ethical Considerations}

The deployment of ML techniques for deception detection carries a number of important ethical implications, especially when applied in sensitive or high-stakes domains such as legal proceedings, border control, clinical settings, or employment screening \parencite{oravec_emergence_2022}. While such systems offer the promise of greater consistency and objectivity compared to human judgment, their use must be carefully evaluated in light of both technical limitations and broader societal concerns.

Under the General Data Protection Regulation (GDPR) in the European Union, individuals are entitled to a "right to explanation" when subjected to decisions made by automated systems that significantly affect them \parencite{goodman2017}. Since many high-performing ML models, such as those used in our study, are difficult to interpret, their use in legally consequential contexts would require robust explainability techniques. This poses both a technical and legal challenge, as explanations must be meaningful to end-users and stakeholders (e.g., defendants, lawyers, clinicians), and not just technically accurate.

Another potential issue that needs to be considered is that, although our approach outperforms untrained human interlocutors in detecting deception, it still falls slightly short of what trained students were able to achieve in one specific study (77.5\% accuracy) \parencite{levine2014diagnostic}, and it is, in general, far from perfect. This creates a risk of misclassification, particularly false negatives (failing to detect a lie) and false positives (incorrectly labeling truthful behavior as deceptive). The consequences of such errors can be significant, ranging from wrongful suspicion or punishment in legal contexts to breakdowns in trust in therapeutic or interpersonal settings. Even with perfect precision in detecting lies (as achieved in our best model), the broader risk profile must be carefully managed before deployment.

An additional concern relates to the potential for bias in the training data or in the model's interpretation of multimodal signals such as facial expressions, tone of voice, or gaze behavior. These cues can vary significantly across cultures, neurodiverse populations, or individuals with different emotional baselines, which could lead to systematic disadvantages for certain groups. This underscores the need for inclusive datasets, fairness audits, and potentially customized models calibrated for specific populations.

Deception detection systems typically rely on rich and sensitive data, including audio, video, and behavioral cues. The use of such data raises privacy concerns, especially in contexts where individuals may not be fully informed or have not consented to being analyzed. Even in research settings, clear ethical guidelines must be followed to protect participants' rights, and any operational use must be aligned with principles of informed consent, data minimization, and secure storage.

Lastly, there is a risk that institutional trust in ML systems could lead to over-reliance on these tools, with human decision-makers deferring to algorithmic outputs even when their limitations are known. This can create a false sense of objectivity, where decisions are viewed as "data-driven" despite the system's probabilistic nature and vulnerability to error.

\subsection{Future Directions}

Future deception detection efforts using multimodal classification approaches, such as the one outlined here, should prioritize generating and maintaining large open-source multimodal datasets for benchmarking and reproducibility. Many research groups have begun the arduous and time-consuming process of systematically collecting multimodal datasets, which is commendable; however, much larger and more diverse samples are required to ensure that classification models are appropriately trained and tested \parencite{delmas_review_2024}, especially if the models are expected to generalize from a specific task to a more general ability to detect deception in a broad array of different situations.

Ideally, such datasets would also account for the distinction between low-stakes lies, which are more easily collected in experimental settings, and high-stakes lies, which are often the focus of practical deception detection efforts employed in law enforcement and criminal justice contexts \parencite{camara_can_2024}. The concern is that, while low-stakes datasets may potentially be more accessible, models trained on such data may not generalize well to high-stakes deception scenarios, where non-verbal behavioral exhibition may differ meaningfully. 

Moreover, future research could also seek to account for individual differences in deception. Previous research has highlighted the seemingly idiosyncratic nature of nonverbal cues in deception contexts, with personality traits, such as extroversion, associated with differing nonverbal behavior exhibition \parencite{zhou_honest_2023}. Furthermore, factors such as childhood development and early social experiences could also be considered, as early exposure to dishonesty during childhood has been linked to a greater tendency to engage in dishonest behavior later in life \parencite{talwar_liar_2022}, potentially further complicating the task of detecting deception across individuals. The use of within-participant experimental designs could enable future research to better account for nonverbal behaviors across truthful and deceptive contexts, although the implementation of such designs may be tricky, as participants could be influenced by order effects and may have difficulty switching between conditions, artificially generating cognitive load which may not accurately reflect the experience of deceivers in real life.

\section{Conclusion} 
Perhaps there is indeed no universal 'Pinocchio's nose'—a single, unmistakable cue signaling deception. Challenges such as the inherent granularity of high-resolution multimodal data, which can introduce significant amounts of noise, as well as individual and cultural differences in expressive behaviors and communication styles, may inherently limit the reliability of deception detection systems \parencite{vrij_reading_2019}. Despite these complexities, this study represents a robust advancement toward more reliable automated detection of deception: by leveraging a novel dataset collected from an underrepresented cohort (native Swedish speakers), this research has systematically compared unimodal and multimodal methods, demonstrating the advantages of multimodal analysis and highlighting the increased predictive power achieved by integrating data from both participants in dyadic interactions. Through this comprehensive  approach, the study underscores the importance and effectiveness of capturing dyadic dynamics rather than relying solely on isolated individual cues. Thus, while perfect detection of deception may remain elusive, this work decisively advances our understanding and methodological capabilities in deception research.

\section{Acknowledgments}
The authors would like to thank Tim Lachmann for his valuable contribution to the preprocessing and extraction of the data used in this study. His support was instrumental to the completion of the data preparation phase.

\section{Declarations}
\subsection{Funding}
This work was supported by the Markus and Amalia Wallembergs Minnesfond under the project ``Låt oss prata om icke-verbal kommunikation" (Let's talk about non-verbal communication).
\subsection{Conflicts of interest}
The authors declare that they have no conflicts of interest.
\subsection{Ethics approval}
No sensitive personal data, as defined by Swedish ethics law, such as information about political views, religious beliefs, or sexual orientation, was collected. All data were pseudo-anonymised using participant codes, with the code key stored securely, separate from the other experimental data. Only members of the research group had access to the data, as originally outlined in the informed consent form. Furthermore, all results are presented on the group level only, further ensuring that no individual participants can be identified. Internal regulations for data collection and handling were followed, including documented approval by the responsible supervisor and the director of studies at the Department of Psychology, Stockholm University.
The research was also approved by the Department of Psychology at Stockholm University.
\subsection{Consent to participate}
Written and verbal informed consent was obtained from all participants. Participants were informed in detail about the study’s purpose and what participation in the study entailed. They were also reminded of their rights, including the right to withdraw at any time without providing a reason. All experimental procedures conformed to the ethical principles outlined in the Declaration of Helsinki (World Medical Association, 2004) and were compliant with the General Data Protection Regulation (GDPR) and Swedish ethical legislation. 
\subsection{Consent for publication}
In the informed consent form, participants were asked to agree to the use of the experimental results for scientific communication. The results are presented at the group level, and no individually identifying information is disclosed.
\subsection{Availability of data and materials}
The authors will make the extracted dataset available upon reasonable request. The shared materials include the features extracted from the original video recordings using OpenFace and OpenSmile, the labels used for the evaluation of the classifiers, and the response data from human participants, which were used to assess participants’ accuracy in judging whether the sender was being truthful or deceptive.
\subsection{Code availability}
All code used for data preprocessing, model training, and analysis is publicly available on GitHub at https://github.com/FoxtrotRomeo/DeceptionDetection
\subsection{Authors' contributions}
Thomas Jack Samuels and Franco Rugolon contributed equally (40\% each) to the conceptualization of the study, the design and execution of the experiments, and the writing of the manuscript.
Stephan Hau and Lennart Högman contributed equally to the supervision, critical review, and revision of the manuscript. All authors read and approved the final version of the paper.

\subsection{Open Practices Statement}
The materials and code used in this study are available on GitHub at https://github.com/FoxtrotRomeo/DeceptionDetection/ to facilitate reproducibility. The dataset used for the analyses is not publicly available due to ethical and privacy constraints, but detailed descriptions of the data collection process are included in the manuscript. This study was not preregistered.

\printbibliography

@article{baltruvsaitis2018multimodal,
  title={Multimodal machine learning: A survey and taxonomy},
  author={Baltru{\v{s}}aitis, Tadas and Ahuja, Chaitanya and Morency, Louis-Philippe},
  journal={IEEE transactions on pattern analysis and machine intelligence},
  volume={41},
  number={2},
  pages={423--443},
  year={2018},
  publisher={IEEE}
}

@inproceedings{zhang2018multi,
  title={Multi-source heterogeneous data fusion},
  author={Zhang, Lili and Xie, Yuxiang and Xidao, Luan and Zhang, Xin},
  booktitle={2018 International conference on artificial intelligence and big data (ICAIBD)},
  pages={47--51},
  year={2018},
  organization={IEEE}
}

@inproceedings{xu2021mufasa,
  title={Mufasa: Multimodal fusion architecture search for electronic health records},
  author={Xu, Zhen and So, David R and Dai, Andrew M},
  booktitle={Proceedings of the AAAI Conference on Artificial Intelligence},
  volume={35},
  number={12},
  pages={10532--10540},
  year={2021}
}

@inproceedings{gadzicki2020early,
  title={Early vs late fusion in multimodal convolutional neural networks},
  author={Gadzicki, Konrad and Khamsehashari, Razieh and Zetzsche, Christoph},
  booktitle={2020 IEEE 23rd international conference on information fusion (FUSION)},
  pages={1--6},
  year={2020},
  organization={IEEE}
}

@inproceedings{kahou2013combining,
  title={Combining modality specific deep neural networks for emotion recognition in video},
  author={Kahou, Samira Ebrahimi and Pal, Christopher and Bouthillier, Xavier and Froumenty, Pierre and G{\"u}l{\c{c}}ehre, {\c{C}}aglar and Memisevic, Roland and Vincent, Pascal and Courville, Aaron and Bengio, Yoshua and Ferrari, Raul Chandias and others},
  booktitle={Proceedings of the 15th ACM on International conference on multimodal interaction},
  pages={543--550},
  year={2013}
}

@article{grossberg1988nonlinear,
  title={Nonlinear neural networks: Principles, mechanisms, and architectures},
  author={Grossberg, Stephen},
  journal={Neural networks},
  volume={1},
  number={1},
  pages={17--61},
  year={1988},
  publisher={Elsevier}
}

@inproceedings{eyben2010opensmile,
  title={Opensmile: the munich versatile and fast open-source audio feature extractor},
  author={Eyben, Florian and W{\"o}llmer, Martin and Schuller, Bj{\"o}rn},
  booktitle={Proceedings of the 18th ACM international conference on Multimedia},
  pages={1459--1462},
  year={2010}
}

@inproceedings{baltruvsaitis2016openface,
  title={Openface: an open source facial behavior analysis toolkit},
  author={Baltru{\v{s}}aitis, Tadas and Robinson, Peter and Morency, Louis-Philippe},
  booktitle={2016 IEEE winter conference on applications of computer vision (WACV)},
  pages={1--10},
  year={2016},
  organization={IEEE}
}

@article{eyben2015geneva,
  title={The Geneva minimalistic acoustic parameter set (GeMAPS) for voice research and affective computing},
  author={Eyben, Florian and Scherer, Klaus R and Schuller, Bj{\"o}rn W and Sundberg, Johan and Andr{\'e}, Elisabeth and Busso, Carlos and Devillers, Laurence Y and Epps, Julien and Laukka, Petri and Narayanan, Shrikanth S and others},
  journal={IEEE transactions on affective computing},
  volume={7},
  number={2},
  pages={190--202},
  year={2015},
  publisher={IEEE}
}

@article{dempster2020rocket,
  title={ROCKET: exceptionally fast and accurate time series classification using random convolutional kernels},
  author={Dempster, Angus and Petitjean, Fran{\c{c}}ois and Webb, Geoffrey I},
  journal={Data Mining and Knowledge Discovery},
  volume={34},
  number={5},
  pages={1454--1495},
  year={2020},
  publisher={Springer}
}

@inproceedings{middlehurst2020canonical,
  title={The canonical interval forest (CIF) classifier for time series classification},
  author={Middlehurst, Matthew and Large, James and Bagnall, Anthony},
  booktitle={2020 IEEE international conference on big data (big data)},
  pages={188--195},
  year={2020},
  organization={IEEE}
}

@article{lee2024z,
  title={Z-Time: efficient and effective interpretable multivariate time series classification},
  author={Lee, Zed and Lindgren, Tony and Papapetrou, Panagiotis},
  journal={Data mining and knowledge discovery},
  volume={38},
  number={1},
  pages={206--236},
  year={2024},
  publisher={Springer}
}

@article{keogh2001dimensionality,
  title={Dimensionality reduction for fast similarity search in large time series databases},
  author={Keogh, Eamonn and Chakrabarti, Kaushik and Pazzani, Michael and Mehrotra, Sharad},
  journal={Knowledge and information Systems},
  volume={3},
  pages={263--286},
  year={2001},
  publisher={Springer}
}

@article{lin2007experiencing,
  title={Experiencing SAX: a novel symbolic representation of time series},
  author={Lin, Jessica and Keogh, Eamonn and Wei, Li and Lonardi, Stefano},
  journal={Data Mining and knowledge discovery},
  volume={15},
  pages={107--144},
  year={2007},
  publisher={Springer}
}

@article{prome2024deception,
  title={Deception detection using ML and DL techniques: A systematic review},
  author={Prome, Shanjita Akter and Ragavan, Neethiahnanthan Ari and Islam, Md Rafiqul and Asirvatham, David and Jegathesan, Anasuya Jegathevi},
  journal={Natural Language Processing Journal},
  pages={100057},
  year={2024},
  publisher={Elsevier}
}

@article{constancio2023deception,
  title={Deception detection with machine learning: A systematic review and statistical analysis},
  author={Const{\^a}ncio, Alex Sebasti{\~a}o and Tsunoda, Denise Fukumi and Silva, Helena de F{\'a}tima Nunes and Silveira, Jocelaine Martins da and Carvalho, Deborah Ribeiro},
  journal={Plos one},
  volume={18},
  number={2},
  pages={e0281323},
  year={2023},
  publisher={Public Library of Science San Francisco, CA USA}
}

@article{tang2018resting,
  title={Resting-state functional connectivity and deception: exploring individualized deceptive propensity by machine learning},
  author={Tang, Honghong and Lu, Xiaping and Cui, Zaixu and Feng, Chunliang and Lin, Qixiang and Cui, Xuegang and Su, Song and Liu, Chao},
  journal={Neuroscience},
  volume={395},
  pages={101--112},
  year={2018},
  publisher={Elsevier}
}

@article{ahmed2024deception,
  title={Deception detection in videos using the facial action coding system},
  author={Ahmed Khan, Hammad Ud Din and Bajwa, Usama Ijaz and Ratyal, Naeem Iqbal and Zhang, Fan and Anwar, Muhammad Waqas},
  journal={Multimedia Tools and Applications},
  pages={1--15},
  year={2024},
  publisher={Springer}
}

@inproceedings{li2013spontaneous,
  title={A spontaneous micro-expression database: Inducement, collection and baseline},
  author={Li, Xiaobai and Pfister, Tomas and Huang, Xiaohua and Zhao, Guoying and Pietik{\"a}inen, Matti},
  booktitle={2013 10th IEEE International Conference and Workshops on Automatic face and gesture recognition (fg)},
  pages={1--6},
  year={2013},
  organization={IEEE}
}

@inproceedings{yan2013casme,
  title={CASME database: A dataset of spontaneous micro-expressions collected from neutralized faces},
  author={Yan, Wen-Jing and Wu, Qi and Liu, Yong-Jin and Wang, Su-Jing and Fu, Xiaolan},
  booktitle={2013 10th IEEE international conference and workshops on automatic face and gesture recognition (FG)},
  pages={1--7},
  year={2013},
  organization={IEEE}
}

@article{yan2014casme,
  title={CASME II: An improved spontaneous micro-expression database and the baseline evaluation},
  author={Yan, Wen-Jing and Li, Xiaobai and Wang, Su-Jing and Zhao, Guoying and Liu, Yong-Jin and Chen, Yu-Hsin and Fu, Xiaolan},
  journal={PloS one},
  volume={9},
  number={1},
  pages={e86041},
  year={2014},
  publisher={Public Library of Science San Francisco, USA}
}

@inproceedings{wan2017results,
  title={Results and analysis of chalearn lap multi-modal isolated and continuous gesture recognition, and real versus fake expressed emotions challenges},
  author={Wan, Jun and Escalera, Sergio and Anbarjafari, Gholamreza and Jair Escalante, Hugo and Bar{\'o}, Xavier and Guyon, Isabelle and Madadi, Meysam and Allik, Juri and Gorbova, Jelena and Lin, Chi and others},
  booktitle={Proceedings of the IEEE international conference on computer vision workshops},
  pages={3189--3197},
  year={2017}
}

@inproceedings{perez2015deception,
  title={Deception detection using real-life trial data},
  author={P{\'e}rez-Rosas, Ver{\'o}nica and Abouelenien, Mohamed and Mihalcea, Rada and Burzo, Mihai},
  booktitle={Proceedings of the 2015 ACM on international conference on multimodal interaction},
  pages={59--66},
  year={2015}
}

@book{ekman2009telling,
  title={Telling lies: Clues to deceit in the marketplace, politics, and marriage (revised edition)},
  author={Ekman, Paul},
  year={2009},
  publisher={WW Norton \& Company}
}

@article{khan2021deception,
  title={Deception in the eyes of deceiver: A computer vision and machine learning based automated deception detection},
  author={Khan, Wasiq and Crockett, Keeley and O'Shea, James and Hussain, Abir and Khan, Bilal M},
  journal={Expert Systems with Applications},
  volume={169},
  pages={114341},
  year={2021},
  publisher={Elsevier}
}

@inproceedings{cardaioli2022face,
  title={Face the truth: interpretable emotion genuineness detection},
  author={Cardaioli, Matteo and Miolla, Alessio and Conti, Mauro and Sartori, Giuseppe and Monaro, Merylin and Scarpazza, Cristina and Navarin, Nicol{\`o}},
  booktitle={2022 International Joint Conference on Neural Networks (IJCNN)},
  pages={01--08},
  year={2022},
  organization={IEEE}
}

@article{poppe2024mining,
  title={Mining bodily cues to deception},
  author={Poppe, Ronald and van der Zee, Sophie and Taylor, Paul J and Anderson, Ross J and Veltkamp, Remco C},
  journal={Journal of Nonverbal Behavior},
  volume={48},
  number={1},
  pages={137--159},
  year={2024},
  publisher={Springer}
}

@inproceedings{gogate2017deep,
  title={Deep learning driven multimodal fusion for automated deception detection},
  author={Gogate, Mandar and Adeel, Ahsan and Hussain, Amir},
  booktitle={2017 IEEE symposium series on computational intelligence (SSCI)},
  pages={1--6},
  year={2017},
  organization={IEEE}
}

@article{loconte2023verbal,
  title={Verbal lie detection using Large Language Models},
  author={Loconte, Riccardo and Russo, Roberto and Capuozzo, Pasquale and Pietrini, Pietro and Sartori, Giuseppe},
  journal={Scientific Reports},
  volume={13},
  number={1},
  pages={22849},
  year={2023},
  publisher={Nature Publishing Group UK London}
}

@article{tsuchiya2023detecting,
  title={Detecting deception using machine learning with facial expressions and pulse rate},
  author={Tsuchiya, Kento and Hatano, Ryo and Nishiyama, Hiroyuki},
  journal={Artificial Life and Robotics},
  volume={28},
  number={3},
  pages={509--519},
  year={2023},
  publisher={Springer}
}

@article{Luke_2019, 
  title={Lessons From Pinocchio: Cues to Deception May Be Highly Exaggerated}, 
  author={Luke, Timothy J.}, 
  journal={Perspectives on Psychological Science}, 
  volume={14}, 
  number={4}, 
  pages={646–671}, 
  year={2019}, 
  doi={10.1177/1745691619838258}, 
  url={https://doi.org/10.1177/1745691619838258} 
}

@article{Matsumoto_Wilson_2023, 
  title={Behavioral Indicators of Deception and Associated Mental States: Scientific Myths and Realities}, 
  author={Matsumoto, David and Wilson, Matthew}, 
  journal={Journal of Nonverbal Behavior}, 
  volume={48}, 
  number={1}, 
  pages={11–23}, 
  year={2023}, 
  month={9}, 
  doi={10.1007/s10919-023-00441-w}, 
  url={https://doi.org/10.1007/s10919-023-00441-w} 
}

@article{Dunbar_Burgoon_Chen_Wang_Ge_Huang_Nunamaker_2023, 
  title={Detecting Ulterior Motives From Verbal Cues in Group Deliberations}, 
  author={Dunbar, Norah E. and Burgoon, Judee K. and Chen, Xunyu and Wang, Xinran and Ge, Saiying and Huang, Qing and Nunamaker, Jay}, 
  journal={Frontiers in Psychology}, 
  volume={14}, 
  year={2023}, 
  month={5}, 
  doi={10.3389/fpsyg.2023.1166225}, 
  url={https://doi.org/10.3389/fpsyg.2023.1166225} 
}

@article{Burgoon_Wang_Chen_Pentland_Dunbar_2021, 
  title={Nonverbal Behaviors “Speak” Relational Messages of Dominance, Trust, and Composure}, 
  author={Burgoon, Judee K. and Wang, Xinran and Chen, Xunyu and Pentland, Steven J. and Dunbar, Norah E.}, 
  journal={Frontiers in Psychology}, 
  volume={12}, 
  year={2021}, 
  month={1}, 
  doi={10.3389/fpsyg.2021.624177}, 
  url={https://doi.org/10.3389/fpsyg.2021.624177} 
}

@article{Brandt_Miller_Hocking_1980b, 
  title={The Truth-Deception Attribution: Effects of Familiarity on the Ability of Observers to Detect Deception}, 
  author={Brandt, David R. and Miller, Gerald R. and Hocking, John E.}, 
  journal={Human Communication Research}, 
  volume={6}, 
  number={2}, 
  pages={99–110}, 
  year={1980}, 
  month={1}, 
  doi={10.1111/j.1468-2958.1980.tb00130.x}, 
  url={https://doi.org/10.1111/j.1468-2958.1980.tb00130.x} 
}

@article{Dunbar_Giles_Bernhold_Adams_2019, 
  title={Strategic Synchrony and Rhythmic Similarity in Lies About Ingroup Affiliation}, 
  author={Dunbar, Norah E. and Giles, Howard and Bernhold, Quinten and Adams, Aubrie and Giles, Matthew and Zamanzadeh, Nicole and Gangi, Katlyn and Coveleski, Samantha and Fujiwara, Ken}, 
  journal={Journal of Nonverbal Behavior}, 
  volume={44}, 
  number={1}, 
  pages={153–172}, 
  year={2019}, 
  month={10}, 
  doi={10.1007/s10919-019-00321-2}, 
  url={https://doi.org/10.1007/s10919-019-00321-2} 
}

@book{vrij_detecting_2000,
	address = {Chichester ; New York},
	series = {Wiley series in the psychology of crime, policing, and law},
	title = {Detecting lies and deceit: the psychology of lying and the implications for professional practice},
	isbn = {978-0-471-85316-9},
	shorttitle = {Detecting lies and deceit},
	publisher = {John Wiley},
	author = {Vrij, Aldert},
	year = {2000},
}

@article{depaulo_lying_1996,
	title = {Lying in everyday life.},
	volume = {70},
	issn = {1939-1315, 0022-3514},
	url = {https://doi.apa.org/doi/10.1037/0022-3514.70.5.979},
	doi = {10.1037/0022-3514.70.5.979},
	language = {en},
	number = {5},
	urldate = {2025-03-11},
	journal = {Journal of Personality and Social Psychology},
	author = {DePaulo, Bella M. and Kashy, Deborah A. and Kirkendol, Susan E. and Wyer, Melissa M. and Epstein, Jennifer A.},
	year = {1996},
	pages = {979--995},
}

@article{fuller_examination_2012,
	title = {An {Examination} of {Deception} in {Virtual} {Teams}: {Effects} of {Deception} on {Task} {Performance}, {Mutuality}, and {Trust}},
	volume = {55},
	copyright = {https://ieeexplore.ieee.org/Xplorehelp/downloads/license-information/IEEE.html},
	issn = {0361-1434, 1558-1500},
	shorttitle = {An {Examination} of {Deception} in {Virtual} {Teams}},
	url = {http://ieeexplore.ieee.org/document/6092522/},
	doi = {10.1109/TPC.2011.2172731},
	number = {1},
	urldate = {2025-03-11},
	journal = {IEEE Transactions on Professional Communication},
	author = {Fuller, Christie M. and Marett, Kent and Twitchell, Douglas P.},
	month = mar,
	year = {2012},
	pages = {20--35},
}

@article{camara_can_2024,
	title = {Can lies be faked? {Comparing} low-stakes and high-stakes deception video datasets from a {Machine} {Learning} perspective},
	volume = {249},
	issn = {09574174},
	shorttitle = {Can lies be faked?},
	url = {https://linkinghub.elsevier.com/retrieve/pii/S0957417424005505},
	doi = {10.1016/j.eswa.2024.123684},
	language = {en},
	urldate = {2025-03-11},
	journal = {Expert Systems with Applications},
	author = {Camara, Mateus Karvat and Postal, Adriana and Maul, Tomas Henrique and Paetzold, Gustavo Henrique},
	month = sep,
	year = {2024},
	pages = {123684},
}

@article{talwar_liar_2022,
	title = {Liar, liar … sometimes: {Understanding} social-environmental influences on the development of lying},
	volume = {47},
	issn = {2352250X},
	shorttitle = {Liar, liar … sometimes},
	url = {https://linkinghub.elsevier.com/retrieve/pii/S2352250X22000938},
	doi = {10.1016/j.copsyc.2022.101374},
	language = {en},
	urldate = {2025-03-11},
	journal = {Current Opinion in Psychology},
	author = {Talwar, Victoria and Crossman, Angela},
	month = oct,
	year = {2022},
	pages = {101374},
}

@article{zloteanu_tutorial_2024,
	title = {A {Tutorial} for {Deception} {Detection} {Analysis} or: {How} {I} {Learned} to {Stop} {Aggregating} {Veracity} {Judgments} and {Embraced} {Signal} {Detection} {Theory} {Mixed} {Models}},
	volume = {48},
	issn = {0191-5886, 1573-3653},
	shorttitle = {A {Tutorial} for {Deception} {Detection} {Analysis} or},
	url = {https://link.springer.com/10.1007/s10919-024-00456-x},
	doi = {10.1007/s10919-024-00456-x},
	language = {en},
	number = {1},
	urldate = {2025-03-11},
	journal = {Journal of Nonverbal Behavior},
	author = {Zloteanu, Mircea and Vuorre, Matti},
	month = mar,
	year = {2024},
	pages = {161--185},
}

@article{levine_accuracy_1999,
	title = {Accuracy in detecting truths and lies: {Documenting} the “veracity effect”},
	volume = {66},
	issn = {0363-7751, 1479-5787},
	shorttitle = {Accuracy in detecting truths and lies},
	url = {http://www.tandfonline.com/doi/abs/10.1080/03637759909376468},
	doi = {10.1080/03637759909376468},
	language = {en},
	number = {2},
	urldate = {2025-03-11},
	journal = {Communication Monographs},
	author = {Levine, Timothy R. and Park, Hee Sun and McCornack, Steven A.},
	month = jun,
	year = {1999},
	pages = {125--144},
}

@article{bond_accuracy_2006,
	title = {Accuracy of {Deception} {Judgments}: {Appendix} {A}},
	volume = {10},
	issn = {1088-8683, 1532-7957},
	shorttitle = {Accuracy of {Deception} {Judgments}},
	url = {http://www.leaonline.com/doi/abs/10.1207/s15327957pspr1003_2A},
	doi = {10.1207/s15327957pspr1003_2A},
	language = {en},
	number = {3},
	urldate = {2025-03-11},
	journal = {Personality and Social Psychology Review},
	author = {Bond, Jr., Charles F. and DePaulo, Bella M.},
	month = jul,
	year = {2006},
}

@article{eskritt_did_2022,
	title = {Did {You} {Just} {Lie} to {Me}? {Deception} {Detection} in {Face} to {Face} versus {Computer} {Mediated} {Communication}},
	volume = {162},
	issn = {0022-4545, 1940-1183},
	shorttitle = {Did {You} {Just} {Lie} to {Me}?},
	url = {https://www.tandfonline.com/doi/full/10.1080/00224545.2021.1933884},
	doi = {10.1080/00224545.2021.1933884},
	language = {en},
	number = {5},
	urldate = {2025-03-11},
	journal = {The Journal of Social Psychology},
	author = {Eskritt, Michelle and Fraser, Brandon and Bosacki, Sandra},
	month = sep,
	year = {2022},
	pages = {566--579},
}

@article{bogaard_no_2022,
	title = {No evidence that instructions to ignore nonverbal cues improve deception detection accuracy},
	volume = {36},
	issn = {0888-4080, 1099-0720},
	url = {https://onlinelibrary.wiley.com/doi/10.1002/acp.3950},
	doi = {10.1002/acp.3950},
	language = {en},
	number = {3},
	urldate = {2025-03-11},
	journal = {Applied Cognitive Psychology},
	author = {Bogaard, Glynis and Meijer, Ewout H.},
	month = may,
	year = {2022},
	pages = {636--647},
}

@article{buller_interpersonal_1996,
	title = {Interpersonal {Deception} {Theory}},
	volume = {6},
	copyright = {http://doi.wiley.com/10.1002/tdm\_license\_1.1},
	issn = {1050-3293, 1468-2885},
	url = {https://academic.oup.com/ct/article/6/3/203-242/4259838},
	doi = {10.1111/j.1468-2885.1996.tb00127.x},
	language = {en},
	number = {3},
	urldate = {2025-03-11},
	journal = {Communication Theory},
	author = {Buller, David B. and Burgoon, Judee K.},
	month = aug,
	year = {1996},
	pages = {203--242},
}

@article{van_der_zee_liar_2021,
	title = {A liar and a copycat: nonverbal coordination increases with lie difficulty},
	volume = {8},
	issn = {2054-5703},
	shorttitle = {A liar and a copycat},
	url = {https://royalsocietypublishing.org/doi/10.1098/rsos.200839},
	doi = {10.1098/rsos.200839},
	language = {en},
	number = {1},
	urldate = {2025-03-11},
	journal = {Royal Society Open Science},
	author = {Van Der Zee, Sophie and Taylor, Paul and Wong, Ruth and Dixon, John and Menacere, Tarek},
	month = jan,
	year = {2021},
	pages = {200839},
}

@article{zhou_honest_2023,
	title = {An {Honest} {Joker} reveals stereotypical beliefs about the face of deception},
	volume = {13},
	issn = {2045-2322},
	url = {https://www.nature.com/articles/s41598-023-43716-4},
	doi = {10.1038/s41598-023-43716-4},
	language = {en},
	number = {1},
	urldate = {2025-03-11},
	journal = {Scientific Reports},
	author = {Zhou, Xingchen and Jenkins, Rob and Zhu, Lei},
	month = oct,
	year = {2023},
	pages = {16649},
}

@article{dunbar_strategic_2020,
	title = {Strategic {Synchrony} and {Rhythmic} {Similarity} in {Lies} {About} {Ingroup} {Affiliation}},
	volume = {44},
	issn = {0191-5886, 1573-3653},
	url = {http://link.springer.com/10.1007/s10919-019-00321-2},
	doi = {10.1007/s10919-019-00321-2},
	language = {en},
	number = {1},
	urldate = {2025-03-11},
	journal = {Journal of Nonverbal Behavior},
	author = {Dunbar, Norah E. and Giles, Howard and Bernhold, Quinten and Adams, Aubrie and Giles, Matthew and Zamanzadeh, Nicole and Gangi, Katlyn and Coveleski, Samantha and Fujiwara, Ken},
	month = mar,
	year = {2020},
	pages = {153--172},
}

@article{vrij_reading_2019,
	title = {Reading {Lies}: {Nonverbal} {Communication} and {Deception}},
	volume = {70},
	issn = {0066-4308, 1545-2085},
	shorttitle = {Reading {Lies}},
	url = {https://www.annualreviews.org/doi/10.1146/annurev-psych-010418-103135},
	doi = {10.1146/annurev-psych-010418-103135},
	language = {en},
	number = {1},
	urldate = {2025-03-11},
	journal = {Annual Review of Psychology},
	author = {Vrij, Aldert and Hartwig, Maria and Granhag, Pär Anders},
	month = jan,
	year = {2019},
	pages = {295--317},
}

@article{colasanti_did_2024,
	title = {Did {You} {Commit} a {Crime} {There}? {Investigating} the {Visual} {Exploration} {Patterns} of {Guilty}, {Innocent}, {Honest}, and {Dishonest} {Subjects} {When} {Viewing} a {Complex} {Mock} {Crime} {Scene}},
	volume = {48},
	issn = {0191-5886, 1573-3653},
	shorttitle = {Did {You} {Commit} a {Crime} {There}?},
	url = {https://link.springer.com/10.1007/s10919-023-00438-5},
	doi = {10.1007/s10919-023-00438-5},
	language = {en},
	number = {1},
	urldate = {2025-03-11},
	journal = {Journal of Nonverbal Behavior},
	author = {Colasanti, Marco and Melis, Giulia and Monaro, Merylin and Ricci, Eleonora and Bosco, Francesca and Rossi, Michela and Biondi, Silvia and Verrocchio, Maria Cristina and Di Domenico, Alberto and Mazza, Cristina and Roma, Paolo},
	month = mar,
	year = {2024},
	pages = {47--71},
}

@article{shen_catching_2021,
	title = {Catching a {Liar} {Through} {Facial} {Expression} of {Fear}},
	volume = {12},
	issn = {1664-1078},
	url = {https://www.frontiersin.org/articles/10.3389/fpsyg.2021.675097/full},
	doi = {10.3389/fpsyg.2021.675097},
	urldate = {2025-03-11},
	journal = {Frontiers in Psychology},
	author = {Shen, Xunbing and Fan, Gaojie and Niu, Caoyuan and Chen, Zhencai},
	month = jun,
	year = {2021},
	pages = {675097},
}

@article{sen_automated_2018,
	title = {Automated {Dyadic} {Data} {Recorder} ({ADDR}) {Framework} and {Analysis} of {Facial} {Cues} in {Deceptive} {Communication}},
	volume = {1},
	issn = {2474-9567},
	url = {https://dl.acm.org/doi/10.1145/3161178},
	doi = {10.1145/3161178},
	language = {en},
	number = {4},
	urldate = {2025-03-11},
	journal = {Proceedings of the ACM on Interactive, Mobile, Wearable and Ubiquitous Technologies},
	author = {Sen, Taylan and Hasan, Md Kamrul and Teicher, Zach and Hoque, Mohammed Ehsan},
	month = jan,
	year = {2018},
	pages = {1--22},
}

@article{matsumoto_clusters_2021,
	title = {Clusters of nonverbal behavior differentiate truths and lies about future malicious intent in checkpoint screening interviews},
	volume = {28},
	issn = {1321-8719, 1934-1687},
	url = {https://www.tandfonline.com/doi/full/10.1080/13218719.2020.1794999},
	doi = {10.1080/13218719.2020.1794999},
	language = {en},
	number = {4},
	urldate = {2025-03-11},
	journal = {Psychiatry, Psychology and Law},
	author = {Matsumoto, David and Hwang, Hyisung C.},
	month = jul,
	year = {2021},
	pages = {463--478},
}

@article{goupil_listeners_2021,
	title = {Listeners’ perceptions of the certainty and honesty of a speaker are associated with a common prosodic signature},
	volume = {12},
	issn = {2041-1723},
	url = {https://www.nature.com/articles/s41467-020-20649-4},
	doi = {10.1038/s41467-020-20649-4},
	language = {en},
	number = {1},
	urldate = {2025-03-11},
	journal = {Nature Communications},
	author = {Goupil, Louise and Ponsot, Emmanuel and Richardson, Daniel and Reyes, Gabriel and Aucouturier, Jean-Julien},
	month = feb,
	year = {2021},
	pages = {861},
}

@article{burgoon_interpersonal_1994,
	title = {Interpersonal deception: {III}. {Effects} of deceit on perceived communication and nonverbal behavior dynamics},
	volume = {18},
	copyright = {http://www.springer.com/tdm},
	issn = {0191-5886, 1573-3653},
	shorttitle = {Interpersonal deception},
	url = {http://link.springer.com/10.1007/BF02170076},
	doi = {10.1007/BF02170076},
	language = {en},
	number = {2},
	urldate = {2025-03-11},
	journal = {Journal of Nonverbal Behavior},
	author = {Burgoon, Judee K. and Buller, David B.},
	month = jun,
	year = {1994},
	pages = {155--184},
}

@article{depaulo_cues_2003,
	title = {Cues to deception.},
	volume = {129},
	issn = {1939-1455, 0033-2909},
	url = {https://doi.apa.org/doi/10.1037/0033-2909.129.1.74},
	doi = {10.1037/0033-2909.129.1.74},
	language = {en},
	number = {1},
	urldate = {2025-03-11},
	journal = {Psychological Bulletin},
	author = {DePaulo, Bella M. and Lindsay, James J. and Malone, Brian E. and Muhlenbruck, Laura and Charlton, Kelly and Cooper, Harris},
	year = {2003},
	pages = {74--118},
}

@article{delmas_review_2024,
	title = {A {Review} of {Automatic} {Lie} {Detection} from {Facial} {Features}},
	volume = {48},
	issn = {0191-5886, 1573-3653},
	url = {https://link.springer.com/10.1007/s10919-024-00451-2},
	doi = {10.1007/s10919-024-00451-2},
	language = {en},
	number = {1},
	urldate = {2025-03-11},
	journal = {Journal of Nonverbal Behavior},
	author = {Delmas, Hugues and Denault, Vincent and Burgoon, Judee K. and Dunbar, Norah E.},
	month = mar,
	year = {2024},
	pages = {93--136},
}

@inproceedings{gupta_bag--lies_2019,
	address = {Long Beach, CA, USA},
	title = {Bag-of-{Lies}: {A} {Multimodal} {Dataset} for {Deception} {Detection}},
	copyright = {https://ieeexplore.ieee.org/Xplorehelp/downloads/license-information/IEEE.html},
	isbn = {978-1-7281-2506-0},
	shorttitle = {Bag-of-{Lies}},
	url = {https://ieeexplore.ieee.org/document/9025340/},
	doi = {10.1109/CVPRW.2019.00016},
	urldate = {2025-03-11},
	booktitle = {2019 {IEEE}/{CVF} {Conference} on {Computer} {Vision} and {Pattern} {Recognition} {Workshops} ({CVPRW})},
	publisher = {IEEE},
	author = {Gupta, Viresh and Agarwal, Mohit and Arora, Manik and Chakraborty, Tanmoy and Singh, Richa and Vatsa, Mayank},
	month = jun,
	year = {2019},
	pages = {83--90},
}

@article{Ekman_Friesen_1969, 
   title={Nonverbal leakage and Clues to deception†}, 
   volume={32}, 
   url={https://doi.org/10.1080/00332747.1969.11023575}, 
   DOI={10.1080/00332747.1969.11023575}, 
   number={1}, 
   journal={Psychiatry}, 
   author={Ekman, Paul and Friesen, Wallace V.}, 
   year={1969}, 
   month=feb, 
   pages={88–106} }

@article{depaulo_nonverbal_1992,
	title = {Nonverbal behavior and self-presentation},
	volume = {111},
	issn = {1939-1455},
	doi = {10.1037/0033-2909.111.2.203},
	number = {2},
	journal = {Psychological Bulletin},
	author = {DePaulo, Bella M.},
	year = {1992},
	note = {Place: US
Publisher: American Psychological Association},
	pages = {203--243},
}

@article{ekman_facial_1978,
	title = {Facial {Action} {Coding} {System}},
	doi = {10.1037/t27734-000},
	author = {Ekman, Paul and Friesen, Wallace V.},
	year = {1978},
}

@article{duran_conversing_2017,
	title = {Conversing with a devil’s advocate: {Interpersonal} coordination in deception and disagreement},
	volume = {12},
	issn = {1932-6203},
	shorttitle = {Conversing with a devil’s advocate},
	url = {https://journals.plos.org/plosone/article?id=10.1371/journal.pone.0178140},
	doi = {10.1371/journal.pone.0178140},
	language = {en},
	number = {6},
	urldate = {2025-03-18},
	journal = {PLOS ONE},
	author = {Duran, Nicholas D. and Fusaroli, Riccardo},
	month = jun,
	year = {2017},
	note = {Publisher: Public Library of Science},
	pages = {e0178140},
}

@article{vrij_why_2004,
	title = {Why professionals fail to catch liars and how they can improve},
	volume = {9},
	copyright = {2004 The British Psychological Society},
	issn = {2044-8333},
	url = {https://onlinelibrary.wiley.com/doi/abs/10.1348/1355325041719356},
	doi = {10.1348/1355325041719356},
	language = {en},
	number = {2},
	urldate = {2025-03-19},
	journal = {Legal and Criminological Psychology},
	author = {Vrij, Aldert},
	year = {2004},
	note = {\_eprint: https://onlinelibrary.wiley.com/doi/pdf/10.1348/1355325041719356},
	pages = {159--181},
}

@article{hartwig_lie_2014,
	title = {Lie {Detection} from {Multiple} {Cues}: {A} {Meta}-analysis},
	volume = {28},
	issn = {1099-0720},
	shorttitle = {Lie {Detection} from {Multiple} {Cues}},
	url = {https://onlinelibrary.wiley.com/doi/abs/10.1002/acp.3052},
	doi = {10.1002/acp.3052},
	language = {en},
	number = {5},
	urldate = {2025-03-19},
	journal = {Applied Cognitive Psychology},
	author = {Hartwig, Maria and Bond Jr., Charles F.},
	year = {2014},
	note = {\_eprint: https://onlinelibrary.wiley.com/doi/pdf/10.1002/acp.3052},
	pages = {661--676},
}

@article{bernieri_dyad_1996,
	title = {Dyad rapport and the accuracy of its judgment across situations: {A} lens model analysis},
	volume = {71},
	issn = {1939-1315},
	shorttitle = {Dyad rapport and the accuracy of its judgment across situations},
	doi = {10.1037/0022-3514.71.1.110},
	number = {1},
	journal = {Journal of Personality and Social Psychology},
	author = {Bernieri, Frank J. and Gillis, John S. and Davis, Janet M. and Grahe, Jon E.},
	year = {1996},
	note = {Place: US
Publisher: American Psychological Association},
	pages = {110--129},
}

@book{han2022data,
  title={Data mining: concepts and techniques},
  author={Han, Jiawei and Pei, Jian and Tong, Hanghang},
  year={2022},
  publisher={Morgan kaufmann}
}

@article{goodman2017,
   title={European Union Regulations on Algorithmic Decision-Making and a “Right to Explanation”},
   volume={38},
   ISSN={0738-4602},
   url={http://dx.doi.org/10.1609/aimag.v38i3.2741},
   DOI={10.1609/aimag.v38i3.2741},
   number={3},
   journal={AI Magazine},
   publisher={Association for the Advancement of Artificial Intelligence (AAAI)},
   author={Goodman, Bryce and Flaxman, Seth},
   year={2017},
   month={10},
   pages={50–57}
}

@article{levine2014diagnostic,
  title={Diagnostic utility: Experimental demonstrations and replications of powerful question effects in high-stakes deception detection},
  author={Levine, Timothy R and Blair, J Pete and Clare, David D},
  journal={Human Communication Research},
  volume={40},
  number={2},
  pages={262--289},
  year={2014},
  publisher={Oxford University Press Oxford, UK}
}

@article{oravec_emergence_2022,
	title = {The emergence of “truth machines”?: {Artificial} intelligence approaches to lie detection},
	volume = {24},
	issn = {1572-8439},
	shorttitle = {The emergence of “truth machines”?},
	url = {https://doi.org/10.1007/s10676-022-09621-6},
	doi = {10.1007/s10676-022-09621-6},
	language = {en},
	number = {1},
	urldate = {2025-04-07},
	journal = {Ethics and Information Technology},
	author = {Oravec, Jo Ann},
	month = {1},
	year = {2022},
	pages = {6},
}

@article{strofer_deceptive_2015,
	title = {Deceptive {Intentions}: {Can} {Cues} to {Deception} {Be} {Measured} before a {Lie} {Is} {Even} {Stated}?},
	volume = {10},
	issn = {1932-6203},
	shorttitle = {Deceptive {Intentions}},
	url = {https://journals.plos.org/plosone/article?id=10.1371/journal.pone.0125237},
	doi = {10.1371/journal.pone.0125237},
	language = {en},
	number = {5},
	urldate = {2025-04-08},
	journal = {PLOS ONE},
	author = {Ströfer, Sabine and Noordzij, Matthijs L. and Ufkes, Elze G. and Giebels, Ellen},
	month = {5},
	year = {2015},
	note = {Publisher: Public Library of Science},
	pages = {e0125237},
}

\end{document}